\newif\ifIsJournal 
\IsJournalfalse

\ifIsJournal
    \documentclass{elsarticle}
    \usepackage{times}
\else
    \documentclass{article}
    \usepackage{iclr_like_style, times}
    \iclrfinalcopy 
\fi

\usepackage{amsmath,amsfonts,bm}

\def\eqref#1{equation~\ref{#1}}

\def\1{\bm{1}}

\def\rva{{\mathbf{a}}}

\def\rvd{{\mathbf{d}}}

\def\rvq{{\mathbf{q}}}

\def\rvx{{\mathbf{x}}}
\def\rvy{{\mathbf{y}}}
\def\rvz{{\mathbf{z}}}

\DeclareMathAlphabet{\mathsfit}{\encodingdefault}{\sfdefault}{m}{sl}
\SetMathAlphabet{\mathsfit}{bold}{\encodingdefault}{\sfdefault}{bx}{n}

\newcommand{\Var}{\mathrm{Var}}

\newcommand{\Ber}{\mathrm{Bernoulli}}

\usepackage{hyperref}
\usepackage{url}

\usepackage{graphicx}
\usepackage{}
\usepackage{booktabs}
\usepackage[symbol]{footmisc}

\usepackage{amssymb}
\usepackage{pifont}
\newcommand{\cmark}{\ding{51}}%
\newcommand{\xmark}{\ding{55}}%
\usepackage{graphicx}  
\usepackage{amsmath} 
\usepackage{amssymb}
\usepackage{mathabx}
\usepackage{hyperref}
\usepackage{glossaries}
\usepackage{enumitem}
\usepackage{arydshln}
\usepackage{makecell}
\usepackage{longtable, tabu}
\usepackage{enumitem}
\usepackage{placeins}
\usepackage{array}
\usepackage{multirow}
\usepackage[T1]{fontenc}
\usepackage[font=small,labelfont=bf,tableposition=top]{caption}
\usepackage{etoolbox}
\usepackage[utf8]{inputenc}
\usepackage{float}
\usepackage{wrapfig,lipsum,booktabs}
\usepackage{caption}
\usepackage{enumitem}
\usepackage{blindtext}
\usepackage{array}
\usepackage{bbm}
\usepackage{todonotes}
\usepackage{cleveref}

\usepackage{xcolor}
\usepackage{soul}
\usepackage{multicol}
\usepackage{tabularx,colortbl}

\newcolumntype{L}{>{\ttfamily}l}

\definecolor{COLORA}{rgb}{0.33999999999999997, 0.86, 0.3712}
\definecolor{COLORB}{rgb}{0.33999999999999997, 0.8287999999999999, 0.86}
\definecolor{COLORC}{rgb}{0.8287999999999999, 0.86, 0.33999999999999997}
\definecolor{COLORD}{rgb}{0.86, 0.33999999999999997, 0.8287999999999999}
\definecolor{COLORE}{rgb}{0.3712, 0.33999999999999997, 0.86}
\definecolor{COLORF}{rgb}{0.86, 0.3712, 0.33999999999999997}
\definecolor{COLORG}{rgb}{0.5, 0.5, 0.5}

\definecolor{niceblueshade}{HTML}{0C6DC7}   
\definecolor{niceblue}{HTML}{1F5B93}   
\definecolor{nicered}{HTML}{BE533B}  
\definecolor{nicegreen}{HTML}{54AD72}  
\definecolor{nicegray}{rgb}{0.3, 0.3, 0.3}

\newcommand\pmx[1]{\raisebox{0.1em}{\color{nicegray}\scriptsize{$\pm$#1}}}

\newcommand{\PatternA}{Correct reasoning step}
\newcommand{\PatternB}{Correct recall of knowledge}
\newcommand{\PatternC}{Correct reading comprehension}
\newcommand{\PatternD}{Incorrect reasoning step}
\newcommand{\PatternE}{Incorrect or insufficient knowledge}
\newcommand{\PatternF}{Incorrect reading comprehension}

\newcommand{\bmtf}{{\footnotesize{BM25}} }

\newcommand{\PromptOne}{\textit{Let's think step by step}}
\newcommand{\PromptTwo}{\textit{Let's think step by step like a medical expert}}

\usepackage{xcolor} 
\usepackage{mdframed} 

\definecolor{lightgray}{rgb}{0.9,0.9,0.9}

\newmdenv[backgroundcolor=lightgray, 
          linecolor=lightgray,
          innerleftmargin=10pt, 
          innerrightmargin=10pt, 
          innertopmargin=10pt, 
          innerbottommargin=10pt,
          skipabove=10pt,
          skipbelow=10pt,
          roundcorner=5pt,
          font=\small]{graybox}

\ifIsJournal
    
\begin{document}
    \begin{frontmatter}
\fi

\title{Can Large Language Models Reason about Medical Questions?}

\ifIsJournal
    \author[dtu,fz]{Valentin Liévin\corref{cor1}}
    \ead{valentin.lievin@gmail.com}
    \cortext[cor1]{Corresponding \& lead author}
    \author[imm]{Christoffer Egeberg Hother}
    \ead{christoffer.egeberg.hother@regionh.dk}
    \author[dtu]{Andreas Geert Motzfeldt}
    \ead{andreas@motzfeldt.dk }
    \author[dtu,fz,gen,ku]{Ole Winther}
    \ead{olwi@dtu.dk}
    \affiliation[dtu]{
                organization={Section for Cognitive Systems, Technical University of Denmark},
                addressline={Anker Engelunds Vej 101}, 
                city={Kongens Lyngby},
                postcode={2800}, 
                country={Denmark}
                }
    \affiliation[fz]{
                organization={FindZebra},
                addressline={Rådvadsvej 36}, 
                city={Copenhagen},
                postcode={2400}, 
                country={Denmark}
                }
    \affiliation[imm]{
                organization={Department of Clinical Immunology, Copenhagen University Hospital, Rigshospitalet},
                addressline={Inge Lehmanns Vej 107}, 
                city={Copenhagen},
                postcode={2100}, 
                country={Denmark}
                }
    \affiliation[gen]{
                organization={Center for Genomic Medicine, Copenhagen University Hospital, Rigshospitalet},
                addressline={Ørestads Blvd. 5}, 
                city={Copenhagen},
                postcode={2300}, 
                country={Denmark}
                }
    \affiliation[ku]{
                organization={Bioinformatics Centre, Department of Biology, University of Copenhagen},
                addressline={Ole Maaløes Vej 5}, 
                city={Copenhagen},
                postcode={2200}, 
                country={Denmark}
                }
\else
    \author{Valentin Li\'evin \textsuperscript{1,2,$\dagger$} 
    Christoffer Egeberg Hother\textsuperscript{3}
    Andreas Geert Motzfeldt \textsuperscript{1}
    Ole Winther\textsuperscript{1, 2, 4, 5, $\dagger$}\\
    \textsuperscript{1} Section for Cognitive Systems, Technical University of Denmark, Denmark\\ 
    \textsuperscript{2} FindZebra ApS, Denmark\\ 
    \textsuperscript{3} Department of Clinical Immunology, Rigshospitalet, Copenhagen University Hospital, Denmark \\
    \textsuperscript{4} Center for Genomic Medicine, Rigshospitalet, Copenhagen University Hospital, Denmark \\
    \textsuperscript{5} Bioinformatics Centre, Department of Biology, University of Copenhagen, Denmark\\
    $\dagger$  Corresponding authors \texttt{valentin.lievin@gmail.com}, \texttt{olwi@dtu.dk}
    }
\fi

\ifIsJournal\else
    \begin{document}
    \maketitle
\fi

\begin{abstract}
Although large language models (LLMs) often produce impressive outputs, it remains unclear how they perform in real-world scenarios requiring strong reasoning skills and expert domain knowledge. We set out to investigate whether close- and open-source models (GPT-3.5, LLama-2, etc.) can be applied to answer and reason about difficult real-world-based questions. We focus on three popular medical benchmarks (MedQA-USMLE, MedMCQA, and PubMedQA) and multiple prompting scenarios: Chain-of-Thought (CoT, think step-by-step), few-shot and retrieval augmentation. Based on an expert annotation of the generated CoTs, we found that InstructGPT can often read, reason and recall expert knowledge. Last, by leveraring advances in prompt engineering (few-shot and ensemble methods), we demonstrated that GPT-3.5 not only yields calibrated predictive distributions, but also reaches the passing score on three datasets: MedQA-USMLE 60.2\%, MedMCQA 62.7\% and PubMedQA 78.2\%. Open-source models are closing the gap: Llama-2 70B also passed the MedQA-USMLE with 62.5\% accuracy.
\end{abstract}

\ifIsJournal
    \end{frontmatter}
\fi

\begin{figure}[H]
    \centering
    \caption{Answering a USMLE (US Medical Licensing Examination) question using zero-shot CoT prompting ``\PromptOne'', \cite{Kojima2022-wg} and InstructGPT~\citep{Ouyang2022-mh}. Selected example.}
    \label{fig:cot-prompting-text}
    \includegraphics[width=1\columnwidth]{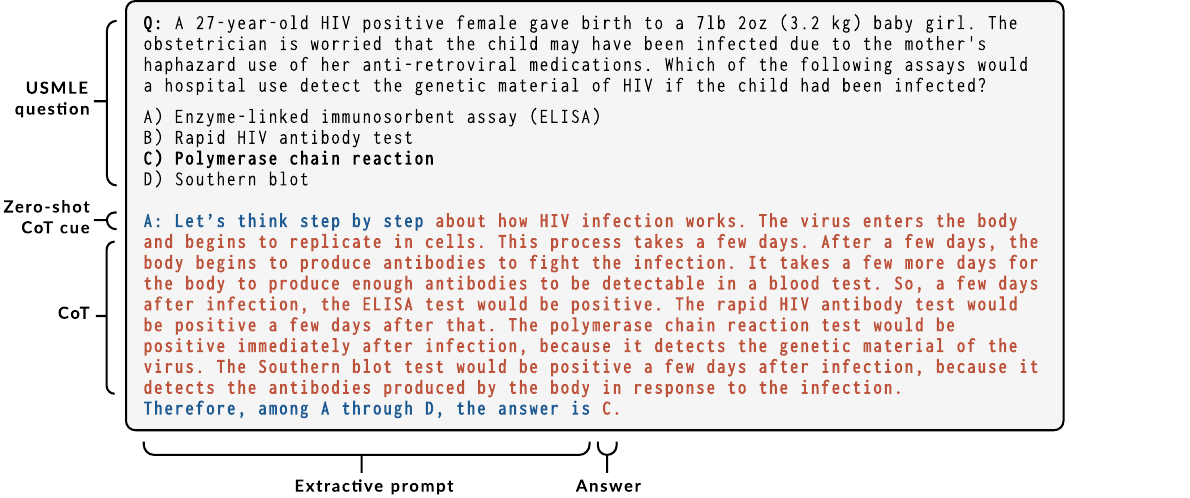}
\end{figure}

\begin{table}[h]
\caption{
Answering accuracy of leading models against human performance on USMLE (test), MedMCQA (validation/test), and PubMedQA (test) datasets. Results marked with $\star$ represent our best methods. 
\ifIsJournal 
Find an overview of our results in Appendix \ref{apdx:master-results}.
\fi}
\label{tab:highlight-table}
\begin{center}
\small
\ifIsJournal
\begin{tabular}{l l l l l}
\toprule
\bf Model & \bf Date & \bf USMLE & \textbf{MedMCQA (dev)} & \bf PubMedQA \\
\midrule
$\star$ Codex 5-shot CoT & \textit{2022} &  60.2 &  59.7 & 78.2 \\
$\star$ Llama-2 5-shot CoT & \textit{2023}  & 62.5 & 53.6 & -- \\
\midrule
Finetuned SOTA  & \textit{2022}  & 50.3 & 52.9 & 78.2 \\
GPT-4 & \textit{2023}  &  86.1 & \textbf{\underline{73.7}} & \textbf{\underline{81.2}} \\
MedPalm v2 & \textit{2023}  &  \textbf{\underline{86.5}} & 72.3 & 77.4 \\
\midrule
Human (passing score)& & 60.0   &  50.0 & -- \\
Human (expert score) & &  87.0  & 90.0 & 78.0 \\
\bottomrule
\end{tabular}
\else
\begin{tabular}{l l l l@{/}l l}
\toprule
\bf Model & \bf Date & \bf USMLE & \multicolumn{2}{c}{\textbf{MedMCQA}} & \bf PubMedQA \\
\midrule
$\star$ Codex 5-shot CoT\textsuperscript{\hyperlink{ht1}{1}} & \textit{2022} &  60.2 &  59.7 & 62.7 & 78.2 \\
$\star$ Llama-2 5-shot CoT\textsuperscript{\hyperlink{ht2}{2}} & \textit{2023}  & 62.5 & 53.6 & -- & -- \\
\midrule
Finetuned SOTA  & \textit{2022}  & 50.3\textsuperscript{\hyperlink{ht3}{3}} & 52.9\textsuperscript{\hyperlink{ht4}{4}} & -- & 78.2\textsuperscript{\hyperlink{ht5}{5}} \\
GPT-4\textsuperscript{\hyperlink{ht6}{6}} & \textit{2023}  &  86.1 & \textbf{\underline{73.7}} &  -- & \textbf{\underline{81.2}} \\
MedPalm v2\textsuperscript{\hyperlink{ht7}{7}} & \textit{2023}  &  \textbf{\underline{86.5}} & 72.3 &  -- & 77.4 \\
\midrule
Human\textsuperscript{\hyperlink{ht8}{8}}  \scriptsize{(passing score)}& & 60.0   &  50.0 & -- & -- \\
Human\textsuperscript{\hyperlink{ht8}{8}} \scriptsize{(expert score)} & &  87.0  & 90.0 & -- & 78.0 \\
\bottomrule
\addlinespace[0.1cm]
\multicolumn{6}{l}{\scriptsize{
$\star$ This paper
,~\hypertarget{ht1}{\textsuperscript{1}}Ensemble of $k$=100 samples, see section \ref{sec:scaling-cot-samples}
,~\hypertarget{ht2}{\textsuperscript{2}}70B parameters, $k$=50 samples.
}}\\
\multicolumn{6}{l}{\scriptsize{
\hypertarget{ht3}{\textsuperscript{3}}PubMedGPT \citep{Venigalla2022-dd}
,~\hypertarget{ht4}{\textsuperscript{4}}Galactica \citep{Taylor2022-ws}
,~\hypertarget{ht5}{\textsuperscript{5}}BioGPT \citep{Luo2022-qh}
}}\\
\multicolumn{6}{l}{\scriptsize{
\hypertarget{ht6}{\textsuperscript{6}}\cite{Nori2023CapabilitiesOG}
,~\hypertarget{ht7}{\textsuperscript{7}}\cite{singhal2023towards}
`~\hypertarget{ht8}{\textsuperscript{8}}See Appendix \ref{apdx:master-results}, Table \ref{tab:master-results}
}}\\
\end{tabular}

\fi
\end{center}
\end{table}

\section{Introduction}

Self-supervised pre-training promises to turn vast quantity of raw data (e.g., text, images, audio) into general-purpose models. Language representations have transformed the field of natural language processing, from simple word vectors \citep{Mikolov2013-gj,Pennington2014-im} to deep contextualized representations \citep{Peters2018-hm, Vaswani2017-st, Devlin2018-qr, Radford2018-kq}, language models are now ubiquitous in natural language processing, notably, thanks to the Transformer architecture \citep{Vaswani2017-st} and its compatibility with massively parallel computation hardware.

\paragraph{Large Language Models (LLMs)} In recent years, tremendous resources have been allocated to scale Transformer-based language models \citep{Brown2020-ad, Rae2021-oy, Chowdhery2022-pr, Thoppilan2022-tm, Hoffmann2022-qq, Smith2022-jc, Zhang2022-zr, Lieber2021-gw, Fedus2021-ci, Laurencon2022-rx} to using hundreds of billions of parameters and to training on gigabytes of text. This so far translated in sustained gains~\citep{Kaplan2020-hw} and enabled new ways to interact with language models. This progress made many of the past benchmarks obsolete and sparked a general interest for designing difficult enough benchmarks (e.g., BIG-bench; \cite{Srivastava2022-jl}). \textit{Pre-train, prompt and predict}~\citep{Liu2021-zx} is an emerging paradigm for applying LLMs to new problems, without fine-tuning the weights on the task. Prompt-based learning consists in augmenting the problem with instructions such that the model's completion of the prompt will correspond to a solution. This allows for LLMs to learn from a few examples (coined \textit{shots}) which are simply incorporated into the prompts~\citep{Brown2020-ad}. 

\paragraph{Chain-of-Thought prompting} Initially, scaling language models up appeared to benefit more knowledge-intensive tasks than the reasoning-heavy ones \citep{Rae2021-oy}. Nevertheless, \cite{Wei2022-tw} demonstrated that LLMs could be applied to \textit{System 2} problems by prompting the model to generate step-by-step solutions, coined ``\textit{Chain-of-Thought}'' (CoT). CoT prompting led to substantial improvements on many reasoning-intensive tasks~\citep{Wei2022-tw, Zhou2022-cd, Drozdov2022-mw, Nye2021-xv}, allowing to bridge the gap with human-level performances for most of the hard BIG-bench tasks~\citep{Suzgun2022-or}. As an alternative to writing reference step-by-step solutions, \textit{zero-shot CoT}~\citep{Kojima2022-wg} allows generating CoTs using single and domain-agnostic cue: ``\PromptOne'' (see example in Figure \ref{fig:cot-prompting-text}). The CoTs that result from that prompt not only appear to expose valid reasoning but also translate into superior zero-shot performances (see example in Figure \ref{fig:cot-prompting-text}).

\paragraph{LLMs and Medical Applications} Applying LLMs to real-life scenarios will require implementing additional safeguards. Language models may amplify the social biases present in the training data, may hallucinate incorrect facts and may lack or robustness~\citep{Bender2021-fv}, for instance to adversarial attacks~\citep{Wang2021-dd}. Therefore, deploying LLMs into sensitive areas such as healthcare must be operated with great care \citep{Korngiebel2021-fc, Sezgin2022-ge}. Nonetheless, large language models are powerful tools and therefore have the potential to transform the field of machine intelligence. At the dawn of this research work, although LLMs had been tested on large benchmarks (MMLU~\cite{Hendrycks2020-jw}, BIG-bench~\cite{Srivastava2022-jl}), studies applied to the medical domain were still needed. Specialized datasets such as the MedQA-USMLE~\citep{Jin2021-jo} enable assessing the capabilities of LLMs in realistic clinical scenarios requiring specialized medical knowledge, advanced reasoning capabilities and human-level reading comprehension skills.

\paragraph{Related Work} This article -- written in three stages (v1: July 2022; v2: December 2022; v3: September 2023) -- evolved along with the remaining of the field. December 2022 was a turning point in machine learning history; new records were achieved on medical benchmarks by the domain-specific Med-PaLM~\citep{Singhal2022LargeLM, singhal2023expertlevel}, ChatGPT\footnote{ChatGPT was released to the public on November 30, 2022 -- \href{https://chat.openai.com}{\texttt{chat.openai.com}}} and GPT-4~\citep{Nori2023CapabilitiesOG}. ChatGPT sparked the interest of the public and the research community, which hastened to benchmark it against USMLE questions \cite{Gilson2022HowDoesChatGPT, Kung2023PerfsOfChatGPT}, turning to self-curated data instead of the peer-reviewed MedQA benchmark.\footnote{USMLE steps 1,2 and 3 were evaluated separately whereas the MedQA aggregates all steps.} Similarly to our work, \cite{Singhal2022LargeLM} and \cite{Kung2023PerfsOfChatGPT} involved human experts to evaluate the generated explanations on USMLE questions. Concurrently, significant progress happened on the open-source world (Llama-2; \cite{touvron2023llama}). Recently, ~\cite{chen2023meditron} investigated both generalist and finetuned open-source LLMs applied to medical benchmarks. CoT prompting and ensemble methods are now commonplace in the literature \citep{Singhal2022LargeLM, singhal2023expertlevel, Nori2023CapabilitiesOG, chen2023meditron} whereas retrieval-augmentation (\textit{grounding}) remains less common \citep{wang2023augmenting, vodqa}.

\paragraph{Contributions}

This paper investigates the performances, interpretability and limitations of CoT prompting for medical question answering. We utilized the GPT-3.5 series (InstructGPT and Codex). This research was conducted in three rounds; first, using InstructGPT, we investigated variations of zero-shot CoT prompting for medical reasoning (domain-specific CoT cues, retrieval augmentation), looking both at the answering performances and the limitations based on an expert evaluation. In the second round, thanks to the Codex beta program, we investigated how scaling inference-time compute could be applied to challenge both the human baseline and to quantify uncertainty. Last, we benchmarked a range of open-source models. Our contributions are:
\begin{itemize}
    \item We assess how GPT-3.5 perform on multiple-choice medical board exam question datasets (MedQA-USMLE and MedMCQA) and a medical reading comprehension dataset (PubMedQA) using prompt engineering. We explore zero-/few-shot, direct/CoT, domain-specific CoT cues and retrieval augmentation.
    \item We propose an evaluation protocol for evaluating generated CoTs (three main categories: reasoning, knowledge and reading comprehension). A medical expert annotated subset of CoTs generated by zero-shot InstructGPT and supports that InstructGPT, in many cases, can reason and exploit memorized expert knowledge.
    \item We demonstrate that scaling inference-time compute enables Codex 5-shot CoT to be well-calibrated and to reach the passing score on the three medical datasets.
    \item We benchmark open-source LLMs on the MedQA-USMLE and MedMCQA.
\end{itemize}

\begin{graybox}
This article has evolved over three distinct versions, each exploring different facets of LLMs:
\vspace{-5pt}
\begin{enumerate}[leftmargin=0.1em]
    \item[] \textbf{v1 - July 2022:} Investigated \textit{InstructGPT} (expert evaluation \& benchmarking prompting strategies).
    \item[] \textbf{v2 - December 2022:} Scaled experiments and passed the MedQA-USMLE using \textit{Codex}.
    \item[] \textbf{v3 - September 2023:} Evaluated open-source models \textit{Llama-2, Vicuna, Guanaco, Falcon, etc.}
\end{enumerate}
\end{graybox}

\section{Method}

\begin{figure}[t]
    \centering
    \caption{Prompt templates. In the table below, we use \texttt{typewriter style} and brackets to represent {\color{niceblue}\texttt{[provided data]}} such as the question, additional context, or the answer and {\color{nicered}\texttt{<completions>}} generated by GPT-3. The symbol $\emptyset$ represents an empty string.}
    \label{fig:prompt-design}
    \includegraphics[width=1.0\columnwidth]{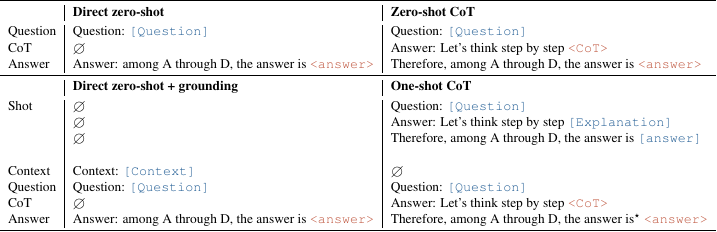}
\end{figure}

This paper explores variations of prompt engineering for medical question answering. The prompt templates are summarized in Figure \ref{fig:prompt-design}.

\paragraph{Zero-shot} We studied two classes of prompts: the \textit{direct} prompt and zero-shot CoT. The direct prompt triggers the model to generate the answer using a single completion step (i.e., ``\textit{The answer is}'') whereas, when applying the zero-shot CoT framework, we use a two-steps prompting scheme: first an initial reasoning prompt with a CoT cue (e.g., ``\PromptOne'') which completion is the CoT, second an extractive prompt which completion is the answer (e.g., ``\textit{Therefore the answer is}''). In the zero-shot CoT setting, this corresponds to the setup described in \cite{Kojima2022-wg}, the direct setting corresponds to \cite{Brown2020-ad}.

\paragraph{Few-shot}

We experimented with inserting examplars (or \textit{shots}) of question-answer pairs and question-explanation-answers triplets in the prompts. We built each shot using the zero-shot template, replacing the output with the reference explanations and answers. In the few-shot CoT setting, our setup matches the one from \cite{Wei2022-tw}.

\begin{figure}[H]
    \centering
    \caption{Generative process and answer likelihood (ensemble model, i.e., self-consistency).}
    \label{fig:cot-illustration}
    \includegraphics[width=0.7\linewidth]{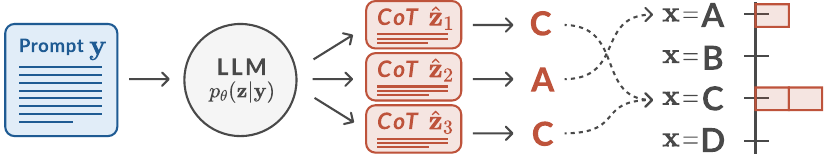}
\end{figure}

\paragraph{Answer likelihood}

We denote $\rvx$ the answer string, $\rvy$ a prompt and $\rvz$ a completion generated from an LLM denoted $p_\theta$. In the zero-shot setting, sampling $\hat{\rvz} \sim p_\theta(\rvz|\rvy)$ is a two-steps process (first generate the CoT, then extract the answer) pictured in Table \ref{tab:prompt-design}. Using a sampling temperature $\tau$, $k$ completions $\hat{\rvz}_1, \ldots, \hat{\rvz}_k$ can be sampled from the generative LLMs. Following \cite{Wang2022-jx}, we aggregate the completions and estimate the marginal answer likelihood as (Figure \ref{fig:cot-illustration})
\begin{equation}\label{eq:likelihood}
    p_\theta(\rvx | \rvy) \approx \frac{1}{k} \sum_{i=1}^k \mathbbm{1}\left[\rvx \in \hat{\rvz}_i\right],\quad \hat{\rvz}_1, \ldots, \hat{\rvz}_k \sim p_\theta(\rvz | \rvy)
\end{equation}
where $\mathbbm{1}\left[\rvx \in \hat{\rvz}_i\right]$ takes value one when the answer $\rvx$ can be matched in the completion $\hat{\rvz}$, otherwise zero. Sampling multiple completions may allow exploring multiple hypotheses. \cite{Wang2022-jx, Li2022-sl} also explored combining multiple sampled CoTs (also known as \textit{self-consistency}) and demonstrated improvements over single-sample CoT methods.

\paragraph{Retrieval augmentation}

LLMs memorise part of the knowledge embedded into the training data, nonetheless, models might fail to re-use this knowledge effectively during prediction. Conditioning the predictions on a knowledge base is an alternative research direction for improving language models \citep{Lewis2020-cg, Borgeaud2021-td, Lazaridou2022-jj}.

We investigated whether \textit{grounding} the model with additional context could improve the answering accuracy. We experimented with a simple \bmtf retriever and used Wikipedia as a knowledge base. Read more details in Appendix \ref{apdx:information-retrieval}.

\section{Experiments}

\begin{table}[h]
\caption{Summary of the medical question answering datasets.}
\label{tab:datasets}
\vspace{-1em}
\begin{center}
\resizebox{\columnwidth}{!}{%
\begin{tabular}{lccc}
\toprule 
 & \bf MedQA-USMLE & \bf MedMCQA &\bf PubMedQA \\
\midrule
Answer options  & A/B/C/D & A/B/C/D & yes/no/maybe  \\
Questions (train/valid./test) & 10.2k/1.3k/1.3k & 182.8k/4.2k/6.1k & 450/50/500   \\
Words / question & 116.6  & 12.7 &  253.3  \\
Source (questions)
& \multicolumn{1}{p{0.25\linewidth}}{\centering \small National Medical Board \\Examination (US)} 
& \multicolumn{1}{p{0.25\linewidth}}{\centering \small AIIMS and NEET PG \\entrance exams} 
& \multicolumn{1}{p{0.25\linewidth}}{\centering \small Expert-annotated \\PubMed abstracts} 
\\
\midrule
Words / explanation & 41.6 & 66.2 & 43.2 \\
Source (explanations) 
& \multicolumn{1}{p{0.25\linewidth}}{\centering \small 5 human-written CoTs \\ \citep{Chung2022-lf}} 
& \multicolumn{1}{p{0.25\linewidth}}{\centering \small Detailed explanations \\(provided)} 
& \multicolumn{1}{p{0.25\linewidth}}{\centering \small Long answer\\ (provided)} 
\\
\bottomrule
\end{tabular}%
}
\end{center}
\end{table}

This section is separated into three parts: (i) introducing the datasets and the GPT-3.5 models, (ii) investigating zero-shot medical reasoning with InstructGPT and (iii) scaling inference-time compute with Codex (using longer few-shot prompts and sampling many completions per question).

\ifIsJournal
    \paragraph{Lead contact}
    Further information and requests for code and data should be directed to and will be fulfilled by the lead contact, Valentin Liévin (valentin.lievin@gmail.com).
    \paragraph{Materials Availability}
    This study did not generate new unique materials or reagents.
    \paragraph{Data and Code Availability}
\fi
Our source code is available on Github.\footnote{\href{https://github.com/vlievin/medical-reasoning}{\texttt{github.com/vlievin/medical-reasoning}} -- \href{https://zenodo.org/doi/10.5281/zenodo.10301873}{DOI: 10.5281/zenodo.10301874}} A collection of generated CoTs, reusable for downstream tasks, are accessible through ToughtSource \citep{ott2023thoughtsource}.\footnote{\href{https://github.com/OpenBioLink/ThoughtSource}{\texttt{github.com/OpenBioLink/ThoughtSource}}} All our benchmark results are summarized in Appendix \ref{apdx:master-results}, Table \ref{tab:master-results}.

\subsection{Datasets and Models}\label{sec:datasets-and-models}

\paragraph{Datasets} 

This study is centered around three medical multiple-choice question answering datasets: USMLE~\citep{Jin2021-jo} which includes difficult real-world medical questions targeting medical professionals, the MedMCQA~\citep{Pal2022-ph} which gathers questions from medical school entrance exams and the PubMedQA~\citep{Jin2019-qa} which includes reading comprehension questions about PubMed abstracts. The three datasets are summarized in Table \ref{tab:datasets}. For each dataset, we gathered questions with explanations (long answer) which we used as reference CoTs in few-shot learning scenarios. We present the three datasets in further details in Appendix \ref{apdx:datasets}. Furthermore, we compare the MedQA-USMLE with the MMLU-USMLE dataset~\citep{Hendrycks2020-jw} in Appendix \ref{apdx:medqa-vs-mmlu}, we found the MedQA questions to be more challenging than the MMLU ones.

\paragraph{Models}\label{sec:models}

We study a collection of closed- and open-source models. The 175B parameter GPT-3.5 series~\citep{Brown2020-ad} the human-aligned GPT-3 (InstructGPT, {\texttt{text-davinci-002}}, \cite{Ouyang2022-mh}), the code-finetuned GPT-3 (Codex, {\texttt{code-davinci-002}}, \cite{Chen2021-sj}). A collection of open-source models ranging from 7B to 70B parameters: Llama-2~\citep{touvron2023llama}, Vicuna~\citep{zheng2023judging}, Guanaco~\citep{dettmers2023qlora}, Falcon~\citep{falcon40b}, MPT~\citep{MosaicML2023Introducing} and GPT-NeoX~\citep{GPT-NeoX-20B}. We used greedy decoding (temperature $\tau=0$) with $k=1$ sample unless specified (e.g., ensemble methods).

In Appendix \ref{apdx:gpt-versions-comparison}, we report the test USMLE accuracy for four GPT-3 versions: a small GPT-3, the largest GPT-3 trained without human-alignment, InstructGPT and Codex. The smaller model {\texttt{text-curie-002}} delivered close to random performances, with a maximum accuracy of 27.9\%. The non-aligned largest GPT-3 {\texttt{text-davinci-001}} scored 40.2\%, whereas the largest code pre-trained Codex scored 52.9\% and the code pretrained and human-aligned InstructGPT scored 47.1\%.

\subsection{Investigating zero-shot reasoning with InstructGPT}\label{sec:exp-instructgpt}

\begin{table}[h]
\caption{Selected domain-specific CoT cues.}
\label{tab:selected-prompts}
\begin{center}
\vspace{-1em}
\begin{tabular}{l}
\toprule 
\footnotesize {\#1} --  \it Let's think step by step \\
\footnotesize {\#2} -- \it Let's think step by step like a medical expert \\
\footnotesize {\#3} -- \it Let’s use step by step inductive reasoning, given the medical nature of the question \\
\footnotesize {\#4} -- \it Let’s differentiate using step by step reasoning like a medical expert\\
\footnotesize {\#5} -- \it Let's derive the differential diagnosis \\
\bottomrule
\end{tabular}%
\end{center}
\end{table}

In this section, we investigate whether the good generative capabilities of LLMs can be applied to answer medical questions in a zero-shot setting. We investigate variations of the zero-shot CoT framework: using domain-specific CoT cues and augmenting the prompt with Wikipedia passages.

\paragraph{Domain-specific CoT prompts} 

In addition to the original zero-shot CoT cue ``\textit{Let's think step by step}'' we tested 29 other domain-specific variations such as ``\textit{Let's think step by step like a medical expert}''. The study is available in Appendix \ref{apdx:prompt-selection}. We selected five CoT cues displayed in Table \ref{tab:selected-prompts}. In Appendix \ref{apdx:additional-CoT-samples}, we display CoT samples for more exotic cues such as ``\textit{Let’s follow a Bayesian step by step approach}'' and ``\textit{Let’s work by elimination}'' and ``\textit{Let’s reflect on each answer option}''.

\subsubsection*{Zero-shot benchmark}\label{sec:zero-benchmark}

\begin{table}[h]
\caption{Zero-shot answering accuracy of InstructGPT (\texttt{text-davinci-002}) on the MedQA-USMLE (test), MedMCQA (valid.) and PubMedQA (test) datasets. We report the best finetuned BERT-based methods. We tested 5 domain-specific CoT cues (\#1-5) and report the mean performances with standard deviations.\ifIsJournal BioLinkBERT: \cite{Yasunaga2022-gl}, PubMedBERT: \cite{Gu2021-jn}, DPR: \cite{Karpukhin2020-di}\fi.
}
\label{tab:zero-shot-benchmark}
\begin{center}
\resizebox{0.9\columnwidth}{!}{%
\ifIsJournal
\begin{tabular}{lcllll}
\toprule
\bf Model & \bf Grounding & \bf Prompt & \bf USMLE & \textbf{MedMCQA} & \bf PubMedQA \\
\midrule
InstructGPT  & \scriptsize{$\emptyset$}       & Direct        &  46.0      &  44.0  &  \textbf{\underline{73.2}} \\
InstructGPT  & \scriptsize{$\emptyset$}       & CoT \#1--5       & 46.1\pmx{0.7}      &  40.4\pmx{2.2}          &  59.9\pmx{3.5} \\
\midrule
InstructGPT  & \scriptsize{BM25}        & Direct        &  47.3      &  46.7           &  -- \\
InstructGPT  & \scriptsize{BM25}         & CoT \#1--5       & 46.4\pmx{0.7}      &  42.5\pmx{1.7}          &  -- \\
\midrule
InstructGPT  & \scriptsize{$\emptyset$}       & Ensemble (n=6)      &  50.0      &  42.4          &  70.4 \\
InstructGPT  & \scriptsize{BM25}        & Ensemble (n=6)      &  49.3      &  \textbf{\underline{48.8}}           &  -- \\
InstructGPT  & \scriptsize{$\emptyset$ + BM25}       & Ensemble (n=12)      &  \textbf{\underline{53.1}}      &  47.6          &  -- \\
\midrule
Finetuned BERT & \scriptsize{BM25, DPR, $\emptyset$} & & 44.6 & 43.0 & 72.2\\ 
\midrule
Human \scriptsize{(passing score)}   & &        & 60.0   & 50.0 & -- \\
Human \scriptsize{(expert score)}    & &    & 87.0 & 90.0 & 78.0 \\
\bottomrule
\end{tabular}
\else
\begin{tabular}{lcllll}
\toprule
\bf Model & \bf Grounding & \bf Prompt & \bf USMLE & \textbf{MedMCQA} & \bf PubMedQA \\
\midrule
InstructGPT  & \scriptsize{$\emptyset$}       & Direct        &  46.0      &  44.0  &  \textbf{\underline{73.2}} \\
InstructGPT  & \scriptsize{$\emptyset$}       & CoT \#1--5       & 46.1\pmx{0.7}      &  40.4\pmx{2.2}          &  59.9\pmx{3.5} \\
\midrule
InstructGPT  & \scriptsize{BM25}        & Direct        &  47.3      &  46.7           &  -- \\
InstructGPT  & \scriptsize{BM25}         & CoT \#1--5       & 46.4\pmx{0.7}      &  42.5\pmx{1.7}          &  -- \\
\midrule
InstructGPT  & \scriptsize{$\emptyset$}       & Ensemble (n=6)\textsuperscript{\hyperlink{vt1}{1}}        &  50.0      &  42.4          &  70.4 \\
InstructGPT  & \scriptsize{BM25}        & Ensemble (n=6)\textsuperscript{\hyperlink{vt1}{1}}        &  49.3      &  \textbf{\underline{48.8}}           &  -- \\
InstructGPT  & \scriptsize{$\emptyset$ + BM25}       & Ensemble (n=12)\textsuperscript{\hyperlink{vt1}{1}}        &  \textbf{\underline{53.1}}      &  47.6          &  -- \\
\midrule
Finetuned BERT & \scriptsize{BM25, DPR\textsuperscript{\hyperlink{vt4}{4}}, $\emptyset$} & & 44.6\textsuperscript{\hyperlink{vt2}{2}} & 43.0\textsuperscript{\hyperlink{vt3}{3}} & 72.2\textsuperscript{\hyperlink{vt2}{2}} \\ 
\midrule
Human \scriptsize{(passing score)}   & &        & 60.0   & 50.0 & -- \\
Human \scriptsize{(expert score)}    & &    & 87.0 & 90.0 & 78.0 \\
\bottomrule
\addlinespace[0.1cm]
\multicolumn{6}{l}{\scriptsize{
\hypertarget{vt1}{\textsuperscript{1}}Majority voting with n predictions, one per prompt (direct + CoT prompts + with/without grounding)
,~\hypertarget{vt2}{\textsuperscript{2}}~BioLinkBERT \citep{Yasunaga2022-gl}}}\\
\multicolumn{6}{l}{\scriptsize{
\hypertarget{vt3}{\textsuperscript{3}}~PubMedBERT \citep{Gu2021-jn} from \cite{Pal2022-ph}.
}
,~\hypertarget{vt4}{\textsuperscript{4}}~DPR \citep{Karpukhin2020-di}}\\
\end{tabular}
\fi
}
\end{center}
\end{table}

In Table \ref{tab:zero-shot-benchmark}, we report the performances of InstructGPT for the direct prompt and the aggregated performances for the five domain-specific CoT cues (Table \ref{tab:selected-prompts}). We explored augmenting the prompts with retrieved Wikipedia passages (grounding) and report the performances of an ensemble model with majority voting, akin to \cite{Wang2022-jx}.

\paragraph{Zero-shot direct} InstructGPT outperformed the domain-specific and finetuned BERT baselines on the three datasets. Without \bmtf grounding, InstructGPT scored +1.4\% on the USMLE questions, +1.0\% on the MedMCQA exam questions and  +1.1\% on PubMedQA over the best BERT methods.

\paragraph{Zero-shot CoT} Without \bmtf grounding, the direct prompt remained, on average, a better alternative to the CoT prompts. Performances were lower for each of the considered CoT cues, except in the case of the USMLE dataset, for which half of the CoT prompts resulted in small improvements over the direct prompt (+1.1\% using the CoT prompt \#1 vs. using the direct prompt). Nonetheless, the domain-specific CoT prompts \#2--5 did not significantly outperform the original CoT prompt \#1.

\paragraph{Knowledge grounding} In an attempt to exploit the good reading comprehension skills of InstructGPT, we explored conditioning the completions on Wikipedia passages. When using the direct prompt, we recorded gains on the USMLE (+1.3\%) and on the MedMCQA (+2.7\%) datasets, suggesting that retrieval augmentation might be beneficial.

\paragraph{Ensemble} Combining the predictions of multiple prompts outperformed the single-prompt predictions, except in the case of the PubMedQA dataset, for which the direct prompt performed exceptionally well. The best performances on the USMLE and MedMCQA datasets were obtained by combining retrieval-augmented prompts, setting a maximum of 53.1\% accuracy on the USMLE dataset and 48.8\% valid. accuracy on the MedMCQA dataset.

\subsubsection*{Expert evaluation of the generated CoTs}\label{sec:patterns}

\begin{table}[H]
\caption{Frequency of observed patterns (A, B, C, D, E, F) identified among 50 CoTs generated by InstructGPT with temperature $\tau$=0. The CoTs are generated based on USMLE questions and using the CoT prompts \#1--5 (Table \ref{tab:selected-prompts}). We report the frequencies of CoTs leading to correct and incorrect predictions along with the total.
}
\label{tab:patterns}
\vspace{-1em}
\begin{center}
\resizebox{0.8\columnwidth}{!}{%
\ifIsJournal
\begin{tabular}{ll@{\hspace{-0.5em}}rrr}
\toprule
{} & \bf Pattern & \bf Correct (16) & \bf Incorrect (34) & \bf Total (50) \\
\midrule
\bf  A & {\PatternA}& 94\% (15) & 59\% (20) & \textbf{70\%} (35)  \\
\addlinespace[0.1cm]
\bf  B & {\PatternB} & 87\% (14)  & 65\% (22) & \textbf{72\%} (36) \\
\addlinespace[0.1cm]
\bf  C & {\PatternC} &  100\% (16)  & 85\% (29) &  \textbf{90\%} (45) \\
\midrule
\bf  D & {\PatternD} & 12\% (2)  & 86\% (29) & \textbf{62\%} (31) \\
\addlinespace[0.1cm]
\bf  E & {\PatternE} & 25\% (4)  & 74\% (25) & \textbf{58\%} (29)  \\
\addlinespace[0.1cm]
\bf  F & {\PatternF} & 6\% (1)  & 50\% (17) & \textbf{36\%} (18) \\
\bottomrule
\end{tabular}
\else
\begin{tabular}{ll@{\hspace{-0.5em}}rrr}
\toprule
{} & \bf Pattern & \bf {\color{niceblue}Correct}\textsuperscript{$\dagger$} (16) & \bf {\color{nicered}Incorrect}\textsuperscript{$\dagger$} (34) & \bf Total (50) \\
\midrule
\bf  A & {\PatternA}\textsuperscript{$\star$} & {\color{niceblue}94\%} (15) & {\color{nicered}59\%} (20) & \textbf{70\%} (35)  \\
\addlinespace[0.1cm]
\bf  B & {\PatternB}\textsuperscript{$\star$} & {\color{niceblue}87\%} (14)  & {\color{nicered}65\%} (22) & \textbf{72\%} (36) \\
\addlinespace[0.1cm]
\bf  C & {\PatternC}\textsuperscript{$\star$} &  {\color{niceblue}100\%} (16)  & {\color{nicered}85\%} (29) &  \textbf{90\%} (45) \\
\midrule
\bf  D & {\PatternD}\textsuperscript{$\star$} & {\color{niceblue}12\%} (2)  & {\color{nicered}86\%} (29) & \textbf{62\%} (31) \\
\addlinespace[0.1cm]
\bf  E & {\PatternE}\textsuperscript{$\star$} & {\color{niceblue}25\%} (4)  & {\color{nicered}74\%} (25) & \textbf{58\%} (29)  \\
\addlinespace[0.1cm]
\bf  F & {\PatternF}\textsuperscript{$\star$} & {\color{niceblue}6\%} (1)  & {\color{nicered}50\%} (17) & \textbf{36\%} (18) \\
\bottomrule
\addlinespace[0.05cm]
\multicolumn{5}{l}{\scriptsize{
\textsuperscript{$\star$}Evidence of (...)
,~\textsuperscript{$\dagger$}For CoTs leading to correct vs. incorrect predictions
}}\\
\end{tabular}
\fi
}
\end{center}
\end{table}

\paragraph{Protocol} InstructGPT delivered strong performances using zero-shot CoT prompting. In this section, we investigate whether the CoTs are sound and seek to understand better how the model fails and succeeds. We considered three general skills that we expect are required to be mastered to answer medical questions: (i) performing non-trivial reasoning steps, (ii) recalling knowledge that is not provided in the context and (iii) the ability to comprehend the question and the context. Based on the three skills, we defined three success patterns (A, B, C) and three failure patterns (D, E, F).

A subset of 50 CoTs generated based on USMLE questions was annotated by a medical expert (C.E.H.) using the six categories. For each category and each CoT, we reported a match if the pattern could be observed at least once. This means that a CoT can be labelled with both a correct and an incorrect pattern for the same skill. We showcase thirty annotated CoTs (three in Figure \ref{fig:usmle-sample-1}, 27 in Appendix \ref{apdx:additional-CoT-samples}). 

\paragraph{Analysis} We report the frequencies of occurrence for the six patterns in Table \ref{tab:patterns}. We found that most of the questions answered incorrectly triggered generating CoTs that contained reasoning errors (pattern D, 86\%), and that exhibited a lack of knowledge (pattern E, 74\%). Misunderstanding of the questions or the context was less frequently observed (Pattern F, 50\%). We observed that CoTs leading to questions answered correctly could still show failure patterns but we also observed that the CoTs leading to incorrect answers were not entirely incorrect, as 59\% contained at least one correct reasoning step, 65\% showed proper recall of knowledge. Furthermore, inspecting the CoTs leading to incorrect answers more closely, we found that 47\% of those were inconclusive:\footnote{Labelling questions as inconclusive or not was also performed by C.E.H.} the model could not narrow down the prediction to a single answer. 

\subsubsection*{Answering bias}\label{sec:bias}
\begin{multicols}{2}
\begin{figure}[H]
\caption{Frequencies of USMLE answers and InstructGPT \texttt{text-davinci-002} predictions for direct and CoT prompts without grounding, zero-shot.}
\label{fig:usmle-gpt3-bias}
\begin{center}
\vspace{-1.0em}
\resizebox{0.8\columnwidth}{!}{%
\includegraphics[width=\columnwidth]{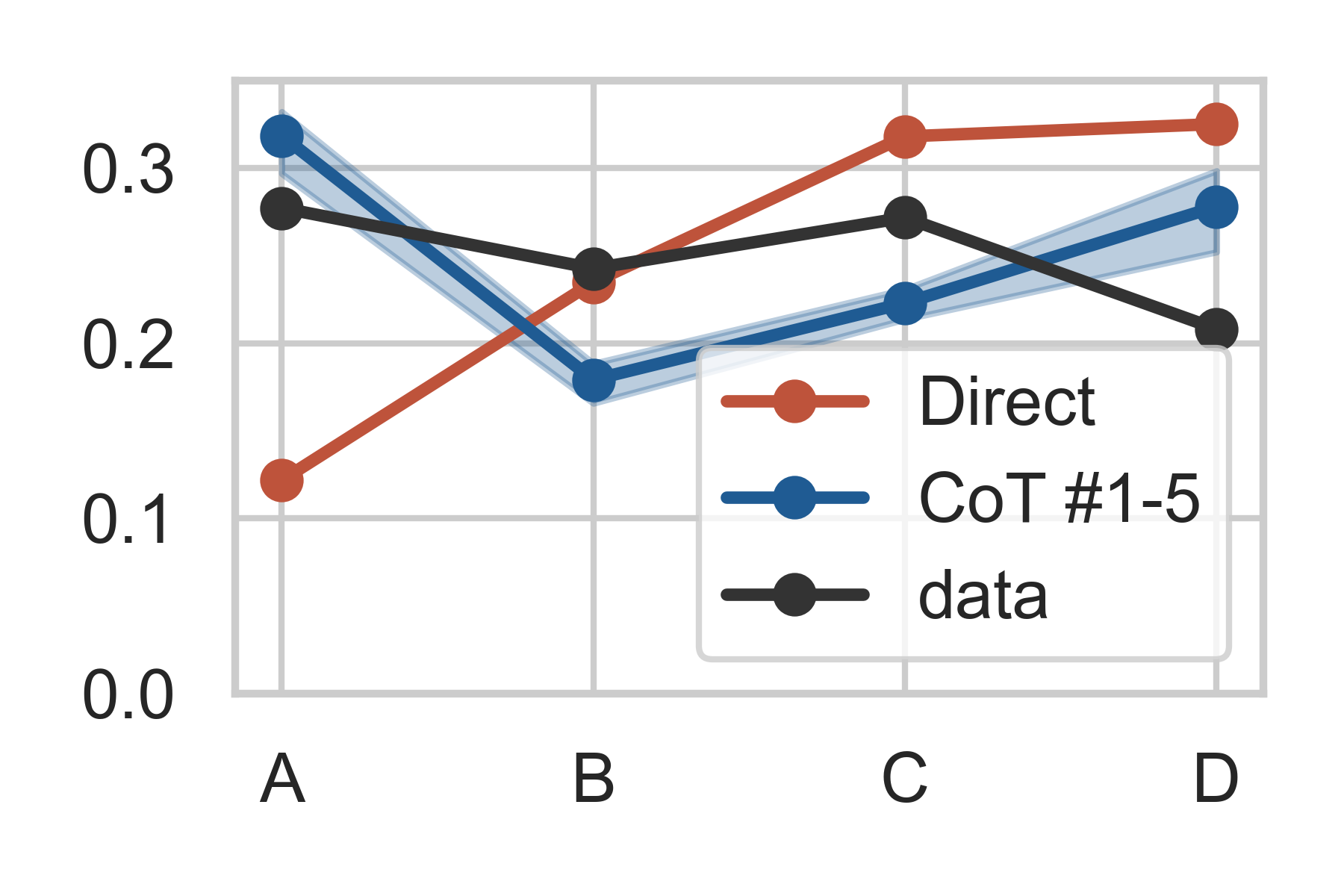}
}
\end{center}
\end{figure}

In Figure \ref{fig:usmle-gpt3-bias}, we report the frequencies of the USMLE answers and the frequencies of predicted labels (zero-shot InstructGPT) for the direct and CoT prompts. Both prompting schemes led to biased predictive frequencies. Direct prompting led to over-estimating the labels C and D while under-estimate the label A. CoT prompting led to under-estimating B and C while over-estimating the label D. We repeat the experiment using randomly permuted labels and observed similar patterns, see Appendix \ref{apdx:answering-bias}.
\end{multicols}

\subsection{Scaling inference-time compute with Codex}\label{sec:exp-codex}

\begin{figure}[h]
    \centering
    \caption{Answering accuracy of Codex 5-shot CoT (\texttt{code-davinci-002}) on the USMLE (test), the MedMCQA (valid.) and the PubMedQA (test) datasets for 100 CoTs sampled with temperature $\tau \in \{0, 0.5\}$. We report the average accuracy for ensemble models evaluated using random subsets of $k'= 1\ldots100$ CoTs. We display the performances of the best finetuned methods along with the lower human baselines.}
    \label{fig:scaling-codex}
    \includegraphics[width=1.0\columnwidth]{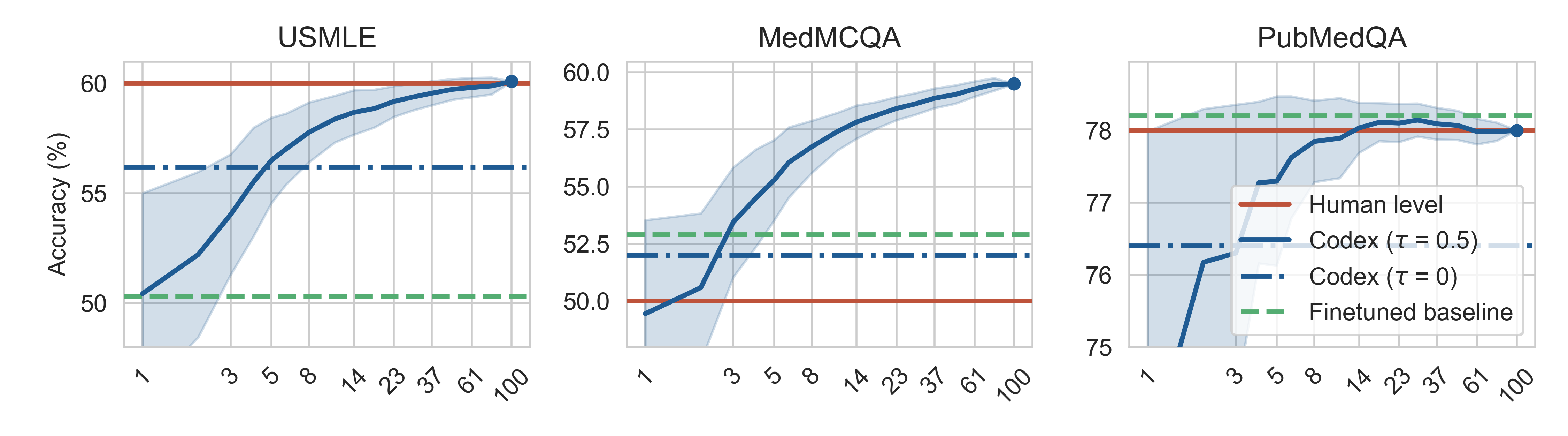}
\end{figure}

In the second round of experiments, we investigated whether using more inference-time compute, thanks to the Codex beta program, could be utilized to obtain better performances and more interpretable outputs. Codex enables using longer prompts, we used five-shot prompts and experimented with sampling $k=100$ completions with temperature $\tau=0.5$ for each question. We report question answering performances and results on uncertainty quantification.

\subsubsection*{Codex 5-shot CoT: sampling and combining multiple CoTs}
\label{sec:scaling-cot-samples}

In Figure \ref{fig:scaling-codex}, we report the performances of Codex 5-shot CoT given subsets of $k' < k$ CoTs. We report the best finetuned models and the human baseline. In line \cite{Wang2022-jx}, increasing the budget of samples yields better results. Using an ensemble of the $k$ samples, Codex 5-shot CoT reaches the passing score on the three tasks (see Table \ref{tab:highlight-table}): the USMLE dataset (60.2\% $\geq$ 60\%), the MedMCQA dataset (62.7\% $\geq$ 50\%) and on the PubMedQA dataset (78.2\% $\geq$ 78\%). Additional results, including performances in zero-shot settings, are available in Table \ref{tab:master-results}, Appendix \ref{apdx:master-results}. Although Codex performed exceptionally well with 5 shots, Codex yield feeble performances with zero-shot CoT; inspecting the generated CoTs revealed lesser-quality samples (Appendix \ref{apdx:additional-CoT-samples}).

\subsubsection*{Uncertainty quantification}
\label{sec:uncertainty}

\begin{figure}[H]
    \centering
    \vspace{-0.2cm}
    \caption{First row: distribution of the probability assigned to the correct label for correct predictions and incorrect predictions (see Equation \ref{eq:likelihood}). Second row: calibration plot. The probabilities are obtained using Codex 5-shot CoT and an ensemble of $k=100$ predictions sampled with temperature $\tau=0.5$.}
    \label{fig:calibration}
    \includegraphics[width=0.9\columnwidth]{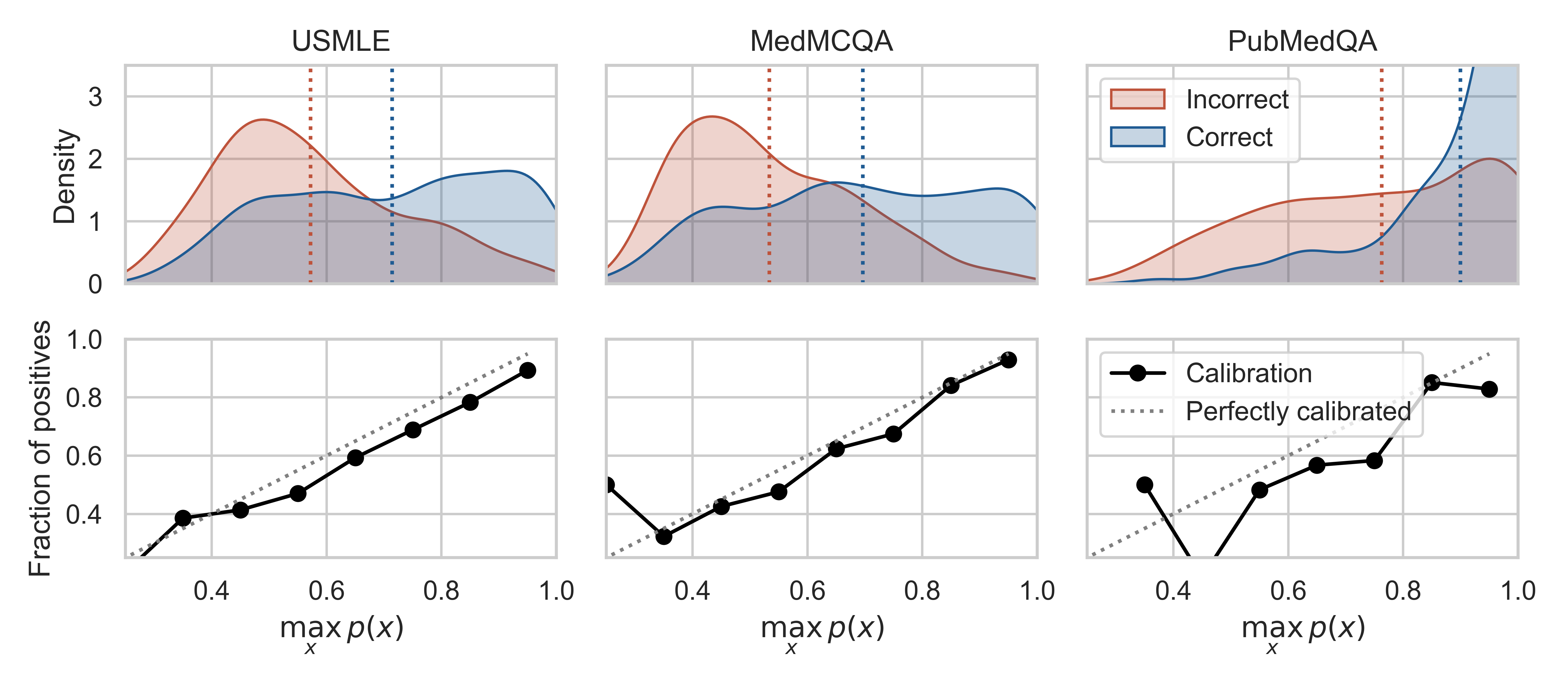}
\end{figure}
\vspace{-0.5cm}
We investigate the answering likelihood Equation \ref{eq:likelihood} given by Codex 5-shot CoT with $k=100$ samples. In Figure \ref{fig:calibration}, we report the maximum probability assigned by the model for correctly vs. incorrectly answered questions along with the calibration plots for the three datasets. Codex 5-shot CoT appears to be overall calibrated, although the calibration is worse for the PubMedQA dataset.

\subsection{Benchmarking Open-Source Models}\label{sec:open-source}

\begin{figure}[t]
    \centering
    \caption{Comparing open-source LLMs against the closed-source Codex on the MedQA-USMLE benchmark ($\tau=0.9$, up to $k=100$ samples). We report answering accuracy, model calibration and answering bias.}
    \label{fig:os-self-consistency}
    \includegraphics[width=\columnwidth]{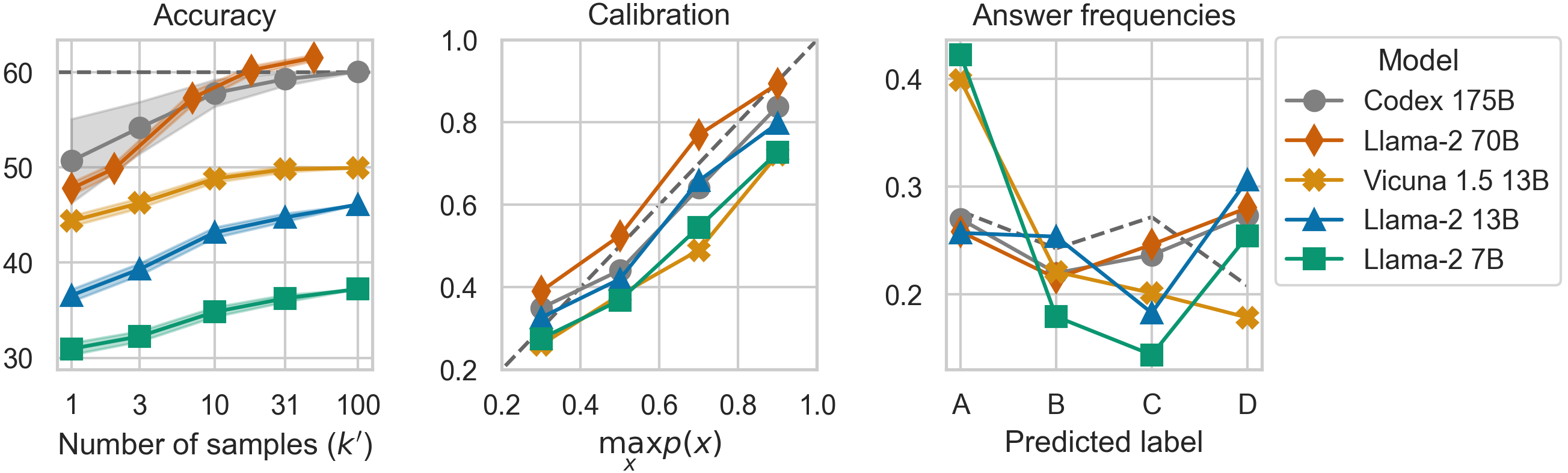}
\end{figure}

In the rapidly evolving landscape of LLMs, a prevalent question is the performance gap between open-source and closed-source models. Our study focused on the capabilities of InstructGPT and Codex. Given a budget of 2.000 A100 hours, we benchmarked a range of open-source LLMs, with parameter sizes ranging from 7 to 70 billion, against the 175-billion-parameter Codex. In Figure \ref{fig:os-self-consistency}, we report the predictive performances, calibration plot and bias for Llama-2, Vicuna 1.5 and Codex using up to $k=100$ CoT samples. We provided additional results in Figure \ref{fig:os-leaderboard} in Appendix \ref{apdx:open-source} (zero- and 5-shot, MedQA-USMLE and MedMCQA).

\section{Discussion}

\paragraph{Zero-shot LLMs outperform finetuned BERT} 

Zero-shot InstructGPT and Codex outperformed finetuned BERT models on three challenging question-answering datasets (section \ref{sec:zero-benchmark} and Appendix \ref{apdx:master-results}). In the case of the USMLE and the MedMCQA datasets, the retrieval-augmented BERT baselines were outperformed by several LLMs, regardless of augmenting the prompts with Wikipedia passages. This suggests that LLMs, without finetuning, can mobilize medical knowledge and problem-solving skills. 

\paragraph{Zero-shot CoT prompting \textit{often} yields sound and interpretable step by step solutions}

For both InstructGPT and Codex, single-sample CoT prompting was not found to be competitive with direct prompting (section \ref{sec:zero-benchmark} and Appendix \ref{apdx:master-results}). Nevertheless, CoTs are human-readable and therefore interpretable. Our expert evaluation (section \ref{sec:patterns}) revealed that CoTs are often sound: even InstructGPT still does mistakes, it was often able to reason, recall medical knowledge and comprehend the given problem. In section \ref{sec:exp-instructgpt} and Appendix \ref{apdx:prompt-selection}, we explored domain-specific CoTs cues such as ``
\PromptTwo''. Although such prompts, taken separately, did not outperform the original zero-shot CoT prompt (see Table \ref{tab:master-results} in Appendix \ref{apdx:master-results}), more specific prompts appeared to trigger alternative strategies such as working by elimination or manipulating equations (see Appendices \ref{apdx:prompt-selection} and \ref{apdx:additional-CoT-samples}). Investigating whether a task-specific prompt could help solve specific tasks will be left for future research. A collection of generated CoT samples are presented in Appendix \ref{apdx:additional-CoT-samples}, many more samples are available on our GitHub page.

\paragraph{LLMs memorize \textit{some} expert knowledge}
The expert evaluation of the generated CoTs (section \ref{sec:patterns}) and the good results obtained on the medical exam questions (see Table \ref{tab:master-results}, Appendix \ref{apdx:master-results}) suggest that GPT-3.5 memorizes domain knowledge. Nevertheless, despite the simplicity of the \bmtf retriever and the small number of retrieved documents prepended in each prompt, grounding InstructGPT resulted in slight improvements (see Table \ref{tab:zero-shot-benchmark}). This suggests that InstructGPT is not omniscient and so (i) using stronger retrievers such as commercial search engines \citep{Lazaridou2022-jj} or dense retrievers \citep{Karpukhin2020-di}, (ii) using a more complete knowledge base \cite{Borgeaud2021-td}, or (iii) leveraging inference-time compute by retrieving, re-ranking and processing more passages \citep{Lazaridou2022-jj} might improve performances. Nevertheless, how to best combine 5-shot CoT prompting with retrieval augmentation remains a promising research direction. 

\paragraph{Bias}

In section \ref{sec:bias}, we exposed the biases induced by the use of direct and CoT prompts. In the case of the direct prompt, the answer D was most often selected, which might be due to its proximity to the generated answer. In the case of the CoT prompts, the labels A and D were selected more often, which might be a result of often beginning CoTs with content related to option A. Based on an inspection of the CoTs, we speculate that GPT-3 defaults to this behaviour when it cannot answer but still attempts to complete the prompt with a default answer (D or A). Shuffling the answer options might be one way to overcome this limitation, however, other forms of biases might still be present.

\paragraph{Generating and combining many CoTs bridges the gap with human-level performances}

CoTs can be combined and/or filtered using human or automated feedback \citep{Wang2022-jx, Cobbe2021-hg}. In section \ref{sec:exp-codex}, we showed that sampling and combining up to $k=100$ completions using Codex or Llama-2 with 5-shot CoT prompts was sufficient to reach both the MedMCQA and the challenging USMLE, although a large gap remains between our models and the human experts.

\paragraph{5-shot CoT-prompted LLMs are close to well-calibrated} In section \ref{sec:uncertainty} and \ref{sec:open-source}, we looked at the probability assigned to correct and incorrect predictions using the ensemble model from Equation \ref{eq:likelihood}. We found Codex and Llama-2 to be close to well-calibrated, corroborating the results of \cite{Kadavath2022-ev} that ``\textit{language models (mostly) know what they know}''.

\paragraph{Scale, code pre-training, human-alignment and few-shot learning}

In Appendix \ref{apdx:gpt-versions-comparison}, we compared multiple GPT-3 models in the zero-shot setting. Best performances are obtained using Codex, outperforming the human-aligned InstructGPT, which is a finetuned version of Codex. Human alignment might impair performances; Codex (without alignment) was not as robust as InstructGPT (with alignment) in zero-shot CoT setting (see performances in Table \ref{tab:master-results} in Appendix \ref{apdx:master-results}, see CoT samples in Appendix \ref{apdx:additional-CoT-samples}). Nevertheless, 5-shot prompting allowed us to bypass the zero-shot limitations of Codex. We observed a similar pattern when comparing the versions of LLama-2 70b: the \texttt{base} version outperformed the \texttt{chat} version (Appendix \ref{apdx:open-source}). Instruction-finetuned models might lose in-context learning abilities. 

\begin{multicols}{2}
\begin{figure}[H]
\caption{MedQA-USMLE accuracy vs. model size. All experiments were performed using a 5-shot CoT prompting strategy and greedy decoding ($\tau=0$). Llama-2 70B outperforms Codex 175B (proprietary).}
\label{fig:os-leaderboard}
\begin{center}
\vspace{-0.9em}
\resizebox{0.9\columnwidth}{!}{%
\includegraphics[width=\columnwidth]{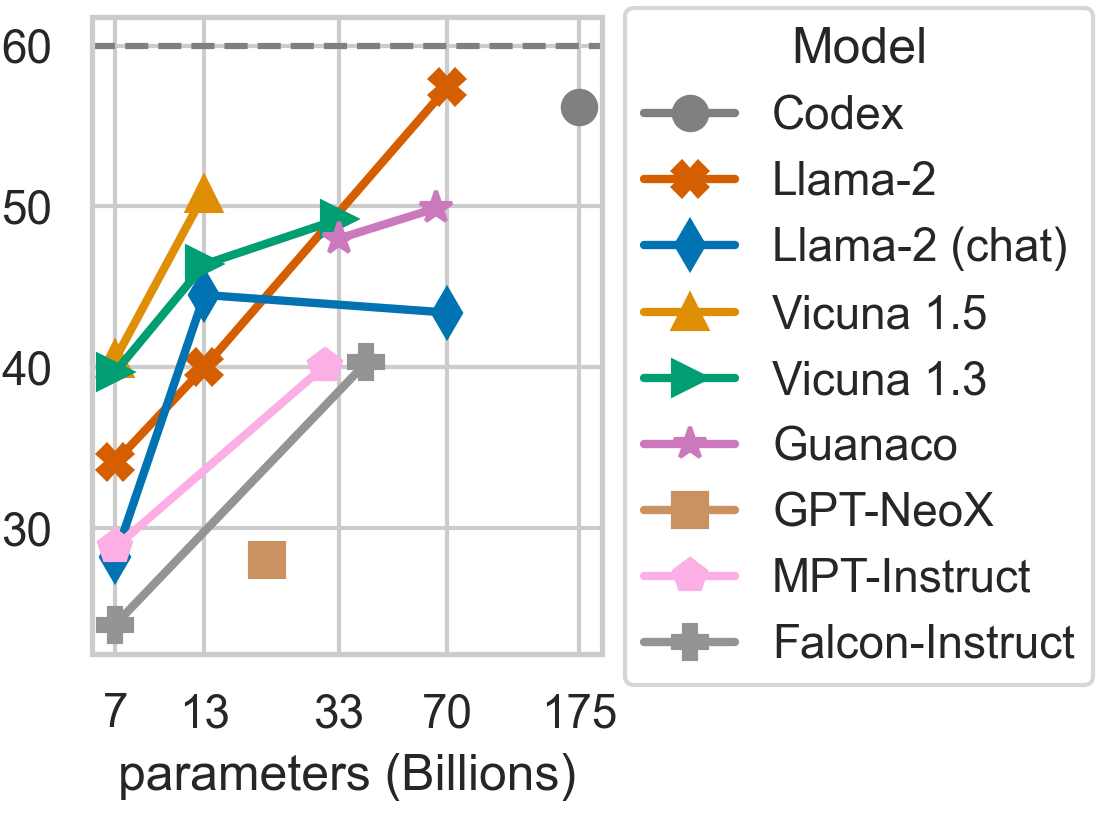}
}
\end{center}
\end{figure}
\columnbreak
\paragraph{Open-source models narrow the gap with proprietary counterparts}

Open-source models, despite having fewer parameters, are approaching the performance of proprietary ones (Figure \ref{fig:os-self-consistency}, \ref{fig:os-leaderboard}). For instance, Llama-2 outperforms Codex with just half the parameters. 

Instruction-finetuned LLMs like Guanaco \citep{dettmers2023qlora} and Vicuna \citep{zheng2023judging} performed exceptionally well (Figure \ref{fig:os-leaderboard}). Surprisingly, Vicuna 1.5 13B's superior performance to both Llama-2 versions underscores the significance of high-quality datasets for instruction-based fine-tuning \citep{zhou2023lima}.
\end{multicols}

\section{Conclusion} 

We applied zero-shot, few-shot direct and CoT prompting to medical question answering with and without retrieval augmentation. Zero-shot InstructGPT significantly outperformed the finetuned BERT baselines. CoT prompting proved to be a powerful tool leading to better performances and more interpretable predictions. Our expert evaluation suggests that, LLMs can mostly comprehend complex medical questions, can often recall expert-domain knowledge and can often perform non-trivial reasoning steps.

Although InstructGPT and Codex still make mistakes, we found that scaling inference-time compute by sampling many CoTs per question could overcome part of these limitations. With 100 samples, Codex 5-shot CoT delivered unprecedented performances on the three datasets, bridging the gap with human-level performances and virtually passing the USMLE by 0.2\% points. Our exploration into open-source LLMs indicated their competitive stance in medical benchmarks. Llama-2 outperformed Codex by 2 points on the USMLE in spite of a much smaller parameter footprint.

However, deploying LLMs in real-life clinical scenarios will require the development of more robust techniques. We exposed one form of bias (ordering of the answer options affects the predictions) but many more might affect predictions, including those hidden in the training data (e.g., gender, race, ...). Nevertheless, a lack of knowledge might be more easily compensated, our experiment with BM25, albeit limited, suggests that augmenting the prompt with factual data improves performances.

Since the completion of version 2 of this work, both GPT-4 and MedPalm 2 have achieved performance on USMLE around 85\% \cite{Nori2023CapabilitiesOG,singhal2023towards}. This is not unexpected given the evolution the LLM field has witnessed recently. Although benchmark contamination in training sets for both proprietary and open-source LLMs is a valid concern, these results indicate that both open and closed sourced LLMs hold great potential for assisting human decision-making in medicine and beyond. 

\subsubsection*{Acknowledgments}
We thank OpenAI for granting access to the Codex beta program. We acknowledge EuroHPC Joint Undertaking for awarding us access to MeluXina at LuxProvide, Luxembourg.
V.L.'s work was funded in part by Google DeepMind through a PhD grant.
OW’s work was funded in part by the Novo Nordisk Foundation through the Center for Basic Machine Learning Research in Life Science (NNF20OC0062606).
V.L., A.G.M. and O.W. acknowledge support from the Pioneer Centre for AI, DNRF grant number P1.

\subsubsection*{Author Contributions}

Conceptualization, V.L., C.E.H. and O.W.; Methodology, V.L. and O.W.; Software, V.L. and A.G.M.; Investigation,
V.L. and A.G.M.; Writing – Original Draft, V.L.; Writing –
Review \& Editing, all authors.; Data Curation, C.E.H. ; Funding Acquisition, O.W.; Supervision, O.W. and V.L.

\ifIsJournal

\subsubsection*{Declaration of Interest}
The authors declare no competing interests.

\subsubsection*{Declaration of Generative AI and AI-assisted technologies in the writing process}
The authors used GPT-3.5 and ChatGPT to help reformulate paragraphs in the writing process. After using generative technologies, the authors reviewed and edited the content as needed and take full responsibility for the content of the publication.
\fi

\clearpage
\begin{figure}[H]
    \centering
    \caption{(Sample 1) Generated zero-shot Chain-of-Thought from InstructGPT \texttt{text-davinci-002} for three CoT prompts on a sample for the MedQA-USMLE test set.}
    \label{fig:usmle-sample-1}
    \includegraphics[width=\columnwidth]{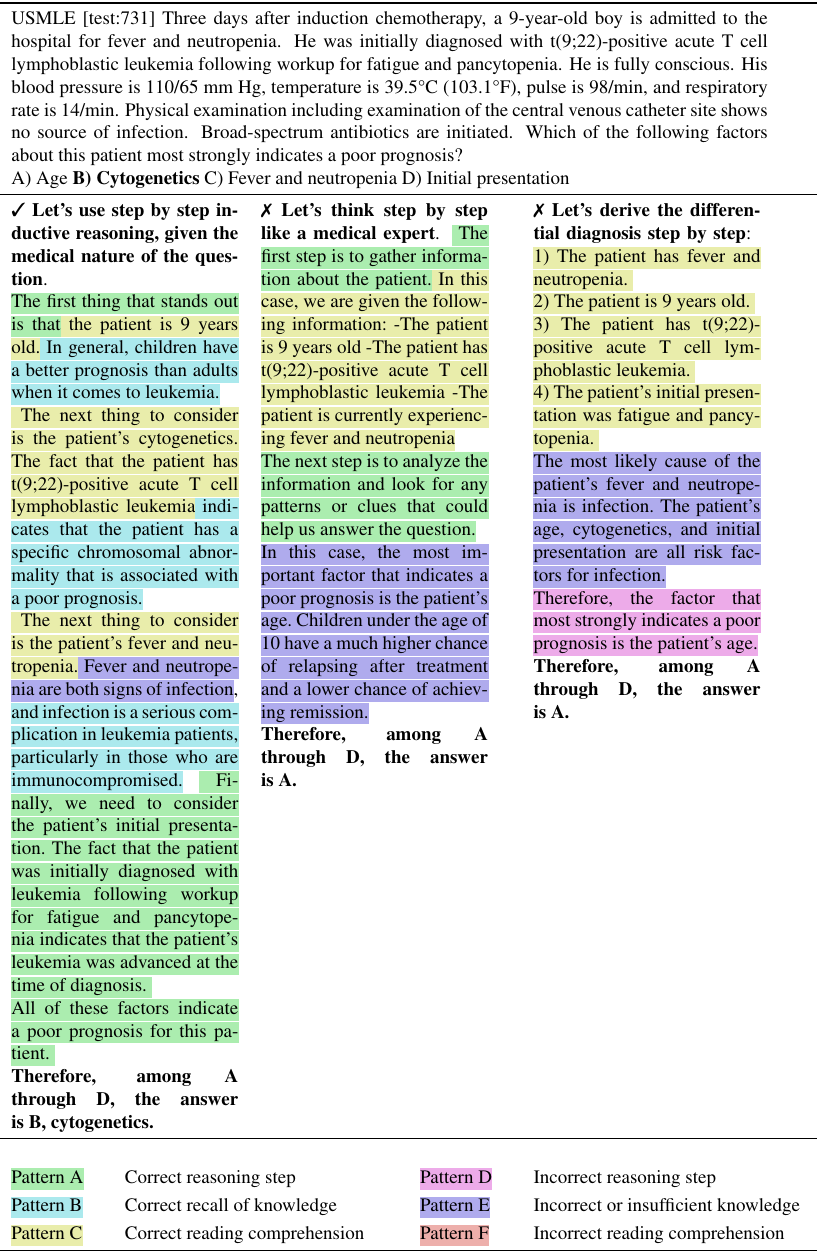}
\end{figure}

\clearpage
\bibliography{references}
\ifIsJournal
    \bibliographystyle{model3-num-names}
\else
    \bibliographystyle{iclr_like_style}
\fi
\clearpage
\appendix

\renewcommand{\thefigure}{S\arabic{figure}}
\renewcommand{\thetable}{S\arabic{table}}
\setcounter{figure}{0}
\setcounter{table}{0}
\pagenumbering{gobble}

\section{Summary of the Results}\label{apdx:master-results}

In Table \ref{tab:master-results}, we summarize the performances of InstructGPT and Codex on the MMLU-USMLE, MedQA-USMLE, MedMCQA and PubMedQA datasets in zero-shot, few-shot, with and without grounding. All our results on the validation set of the MedMCQA are estimated using 1k samples. The results of the MedMCQA test set require submitting an official submission. We used a sampling temperature of $\tau=0$ for all experiments except when drawing $k > 0$ samples and using majority voting (MV). For the majority voting model, we used $k=100$ samples and $\tau=0.5$ for Codex, $\tau=0.9$ for Vicuna.

\section{Domain-specific CoT cues}\label{apdx:prompt-selection}
\begin{table}[h]
\caption{Validation performances for 30 CoT cues on a subset of 100 validation USMLE questions.}
\label{tab:usmle-validation}
\begin{center}
\resizebox{0.9\columnwidth}{!}{%
\begin{tabular}{llrrr}
\toprule
{} & \bf CoT cue & \bf Accuracy &    \bf F1 & \bf CoT length \\
\midrule
0  & \it                                  Let's derive the differential diagnosis step by step &    \textbf{\underline{48.0}} & 48.0 &      170 \\
1  & \it  Let's use step by step inductive reasoning, given the medical nature of the question &    \textbf{\underline{48.0}} & \textbf{\underline{48.2}} &      157 \\
2  & \it                Let's differentiate using step by step reasoning like a medical expert &    47.0 & 46.3 &      183 \\
3  & \it                                    Let's think step by step using deductive reasoning &    47.0 & 46.4 &      148 \\
4  & \it                                      Let's differentiate using step by step reasoning &    45.0 & 45.0 &      166 \\
5  & \it                              Let's think step by step to arrive at one of the options &    45.0 & 45.0 &      158 \\
6  & \it                                           Let's break the problem into multiple steps &    45.0 & 44.2 &      165 \\
7  & \it  Let's use step by step deductive reasoning, given the medical nature of the question &    44.0 & 44.0 &      174 \\
8  & \it                                                Let's think step by step like a doctor &    43.0 & 43.3 &      162 \\
9  & \it                                        Let's think step by step like a medical expert &    43.0 & 42.8 &      171 \\
10 & \it                                                Let's summarize the facts step by step &    42.0 & 42.1 &      183 \\
11 & \it                                    Let's think step by step using inductive reasoning &    42.0 & 42.6 &      143 \\
12 & \it              Let's think step by step using deductive reasoning like a medical expert &    42.0 & 42.3 &      173 \\
13 & \it                                               Let's be concise and think step by step &    42.0 & 42.4 &      130 \\
14 & \it      Let's differentiate using step by step deductive reasoning like a medical expert &    42.0 & 41.9 &      173 \\
15 & \it                                                              Let's argue step by step &    42.0 & 42.2 &      149 \\
16 & \it                                             Let's think step by step like a clinician &    41.0 & 41.3 &      164 \\
17 & \it                                                              Let's think step by step &    40.0 & 40.4 &      129 \\
18 & \it                                      Let's reflect on each answer option step by step &    40.0 & 37.2 &      194 \\
19 & \it             Let's reason and differentiate options step by step like a medical expert &    40.0 & 38.1 &      180 \\
20 & \it      Let's differentiate using step by step inductive reasoning like a medical expert &    40.0 & 39.5 &      161 \\
21 & \it                                                                $\emptyset$ (Direct) &    39.0 & 38.4 &      0 \\
22 & \it                       Let's think step by step given every option equal consideration &    39.0 & 39.2 &      177 \\
23 & \it                                             Let's think step by step like a scientist &    39.0 & 39.2 &      166 \\
24 & \it                                            Let's use step by step inductive reasoning &    37.0 & 36.1 &      165 \\
25 & \it                                                Let's work by elimination step by step &    36.0 & 35.2 &      154 \\
26 & \it                                            Let's use step by step deductive reasoning &    34.0 & 33.9 &      165 \\
27 & \it                                         Let's follow a Bayesian step by step approach &    33.0 & 31.4 &      193 \\
28 & \it                 Let's reflect on each option from the least likely to the most likely &    31.0 & 27.9 &      166 \\
29 & \it   Let's use step by step Bayesian reasoning, given the medical nature of the question &    31.0 & 30.7 &      216 \\
\bottomrule
\end{tabular}

}
\end{center}
\end{table}

We composed an initial set of 30 zero-shot CoT prompt variations. In Table \ref{tab:usmle-validation}, we report the accuracy for each of the 30 prompts based on a subset of 100 USMLE validation questions. Given an estimated accuracy uncertainty of 5\% (see the paragraph ``uncertainty estimation'' below), we concluded that the first half of the results are all reasonable candidates for the study. 

\paragraph{Selected prompts} For the remaining of this paper, we selected 5 prompts: the original ``\textit{Let's think step by step}'', the medical variation ``\textit{Let's think step by step like a medical expert}'' and the top-three CoT cues reported in Table \ref{tab:usmle-validation}.

\begin{figure}[H]
  \centering
  \caption{Rate of agreement for the 30 evaluated CoT prompts evaluated in Table \ref{tab:usmle-validation}.}
  \label{fig:experts-agreement}
    \ifIsJournal
    \includegraphics[width=0.9\textwidth]{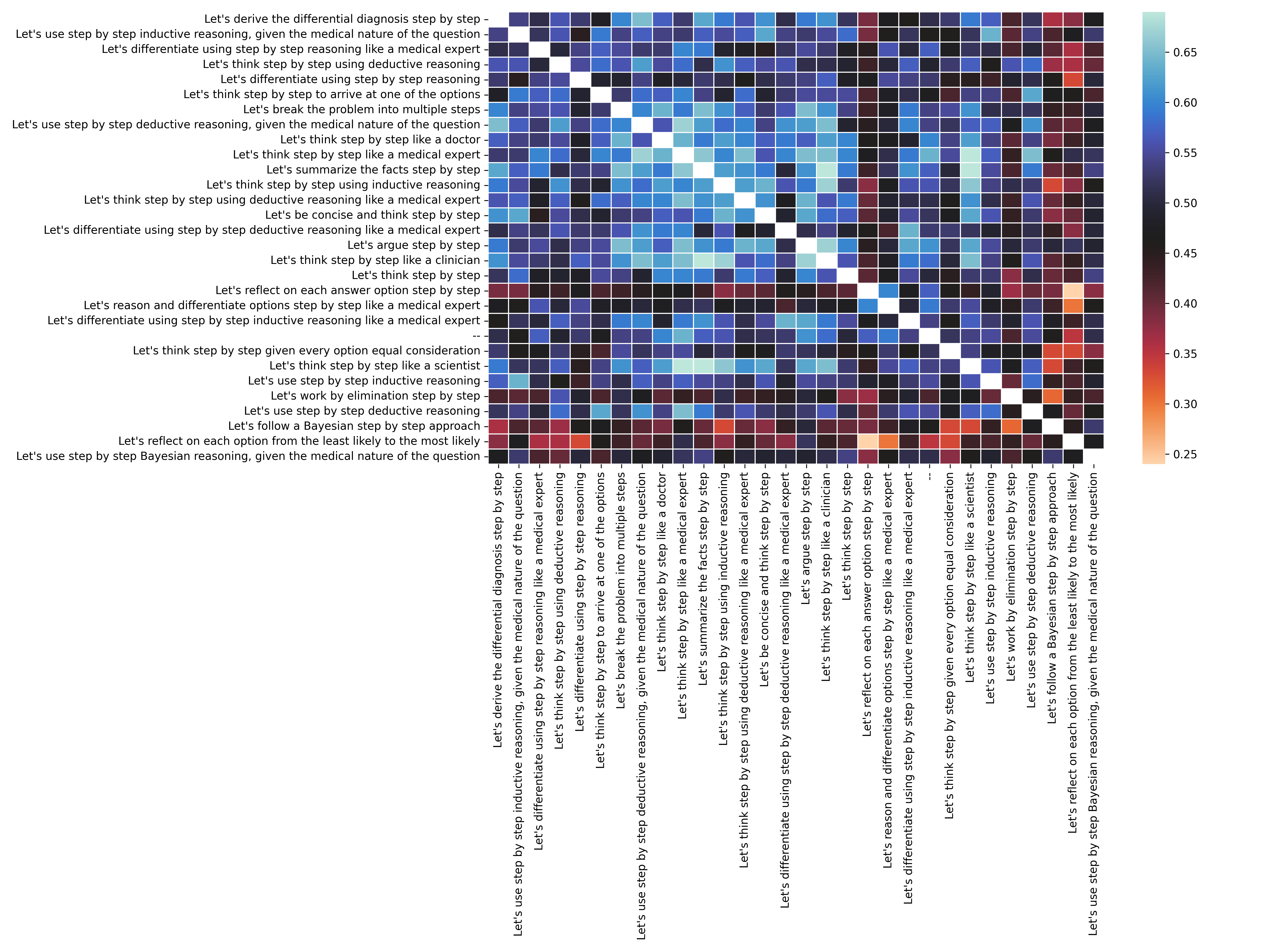}
    \else
    \includegraphics[width=1.1\textwidth]{figures/Figure_S10.png}
    \fi
\end{figure}

\paragraph{Prompt diversity and agreement} In Figure \ref{fig:experts-agreement}, we report the agreement rate for all 30 prompts on the 100 validation questions. Whereas most of the prompts followed a rather consistent pattern, with an agreement rate superior to 50\%, a minority of the prompts seemed to agree less with the majority of the prompts, such as ``Let's reflect on each answer option step by step'', ``Let's follow a Bayesian step by step approach'' or ``Let's work by elimination''. In Figure \ref{fig:usmle-remarkable-strategies}, we showcase four chain-of-thoughts selected to highlight the diversity of the completions and the ability of InstructGPT to adopt diverse problem-solving strategies. Yet, strategies are not always executed correctly: in Figure \ref{fig:usmle-remarkable-strategies}, example 2, GPT-3 ultimately finds the correct answer (Missense mutation) but identified the wrong diagnostic (the 6-year-old boy suffers from sickle cell disease).

\paragraph{Uncertainty estimation}

We model the outcome of answering a question using a $\Ber$ model with parameter $\theta$ where 1 corresponds to the correct predicted answer, and 0 corresponds to predicting the wrong answer. The accuracy of the model corresponds to the mean outcome of the $\Ber$ model ($\mathbb{E}\left[ \Ber(\theta) \right] = \theta$) that we approximate as $\theta=0.5$. Given N=100 data points, the uncertainty of the accuracy estimate is about 5\%, as given by the standard deviation of the mean estimator:
\begin{equation*}
    \sqrt{\Var_N \left[ \Ber(\theta) \right]} = \sqrt{\frac{\theta (1-\theta)}{N} } = 0.5^2 / 100 = 0.05\ (5\%) \ .
\end{equation*}

\ifIsJournal
\else
\vspace{2.0cm}
\fi

\begin{table}[h]
\vspace{-1.2cm}
\begin{center}
\caption{Question answering accuracy on the USMLE, MedMCQA (valid. 1k samples/test), and PubMedQA datasets. The CoTs cues \#1--5 are defined in Table \ref{tab:selected-prompts} (e.g., \#2 = \textit{Let's think step by step like a medical expert}). Results marked with $\star$ represent the pinnacles of our observations.}
\label{tab:master-results}
\resizebox{!}{11.0cm}{%
\begin{tabular}{lccl r@{\hspace{1pt}}l c r@{\hspace{1pt}}l c}
\toprule
\bf Model & \bf Shots & \bf Grounding &\bf Prompt & \multicolumn{2}{c}{\textbf{MMLU}~\textsuperscript{\hyperlink{tt1}{1}}} & \bf USMLE & \multicolumn{2}{c}{\centering \textbf{MedMCQA}} & \bf PubMedQA \\
\midrule
InstructGPT (175B) & 0 & \xmark             & Direct   & -- &/ --       & 46.0      &  44.0 &/ --           &  73.2 \\
InstructGPT (175B) & 0 &  \xmark            & CoT \#1      & -- &/ --       & 47.1      &  40.8 &/ --           &  60.0 \\
InstructGPT (175B) & 0 &  \xmark            & CoT \#2      & -- &/ --       & 46.8      &  43.3 &/ --           &  59.8 \\
InstructGPT (175B) & 0 &  \xmark            & CoT \#3      & -- &/ --       & 46.0      &  38.8 &/ --           &   66.2 \\
InstructGPT (175B)  & 0 &  \xmark           & CoT \#4      & -- &/ --       & 45.6      &  37.1 &/ --           &   58.0 \\
InstructGPT (175B)  & 0 &  \xmark           & CoT \#5      & -- &/ --       & 45.1      &  42.1 &/ --           &   55.6 \\
\addlinespace[0.15cm]
InstructGPT (175B)  & 0 &  \cmark           & Direct   & -- &/ --       & 47.3      &  46.7 &/ 49.0         & -- \\
InstructGPT (175B)  & 0 &  \cmark           & CoT \#1      & -- &/ --       & 45.9      &  42.2 &/ 46.0         &  -- \\
InstructGPT (175B)  & 0 &  \cmark           & CoT \#2      & -- &/ --       & 47.0      &  45.8 &/ 46.0         &  -- \\
InstructGPT (175B)  & 0 &  \cmark           & CoT \#3      & -- &/ --       & 45.6      &  41.6 &/ 43.3         &   -- \\
InstructGPT (175B)  & 0 &  \cmark           & CoT \#4      & -- &/ --       & 45.9      &  41.3 &/ 45.0         &   -- \\
InstructGPT (175B)  & 0 &  \cmark           & CoT \#5      & -- &/ --       & 47.4      &  41.8 &/ 46.5         &   --  \\
\addlinespace[0.15cm]
InstructGPT (175B)  & 0 &  \xmark           &  Ensemble (n=6)  & -- &/ --       &  50.0     &  42.4 &/ --           & 70.4 \\
InstructGPT (175B)  & 0 &  \cmark           &  Ensemble (n=6)  & -- &/ --       & 49.3      & 48.8 &/ 49.5   & --\\
InstructGPT (175B)  & 0 &  \xmark\ + \cmark &  Ensemble (n=12)  & -- &/ --       & 53.1  &  47.6 &/ --           & --\\
\midrule
Codex (external results)\textsuperscript{\hyperlink{tt14}{14}}    & 0 &  \xmark           &  Direct     & -- &/ --       & --        &  -- &/ 54.4             & -- \\
\addlinespace[0.15cm]
Codex (175B)  & 0 &  \xmark           &  Direct  & 74.2 &/ 70.6  & 52.5      &  50.9 &/ 51.1         & 73.2 \\ 
Codex (175B)  & 0 &  \xmark           &  CoT \#1     & 64.5 &/ 60.7   & 52.9      &  46.8 &/ --           & 53.4 \\ 
\addlinespace[0.15cm]
Codex (175B)  & 0 &  \cmark           &  Direct  & 64.5 &/ 68.7    & 52.5      &  50.8 &/ 52.7        & -- \\ 
Codex (175B)  & 0 &  \cmark           &  CoT \#1     & 77.4 &/ 55.1     & 47.2       &  43.9 &/ --          &  -- \\ 
\addlinespace[0.15cm]
Codex (175B)  & 5 &  \xmark           &  Direct  & 77.2 &/ 70.1   & 56.6      &  56.6 &/ 56.9         & 73.0 \\ 
Codex (175B)  & 5 &  \xmark           &  CoT \#1     & $\star$80.6 &/ 68.4   & 56.2     &  52.0 &/ --           & 76.4\\ 
Codex (175B)  & 5 &  \xmark           &  CoT \#1 + MV($k$=20)\textsuperscript{\hyperlink{tt10}{10}}     & 74.2 &/ $\star$\underline{\textbf{76.8}}   
& 57.2    &  57.6 &/ 57.5            & $\star$78.2 \\ %
Codex (175B)  & 5 &  \xmark           &  CoT \#1 + MV($k$=100)\textsuperscript{\hyperlink{tt10}{10}}     & -- &/ --   
& 60.2      &  $\star$59.7 &/ $\star$62.7          & 78.0 \\
\midrule
GPT-4   & 5 &  \xmark           &  CoT \textsuperscript{\hyperlink{tt21}{21}}     & -- &           &  86.1  & \underline{\textbf{73.7}} & / --           &   77.4 \\   
\midrule
GPT-NeoX (20B)\textsuperscript{\hyperlink{tt20}{20}}        & 0 &  \xmark           & Direct      & -- &           &  26.9 & 27.8 & / --            &   -- \\
GPT-NeoX (20B)        & 5 &  \xmark           & CoT \#1      & -- &           &  28.0 & 32.9 & / --           &   -- \\
\addlinespace[0.15cm]
MPT-Instruct (7B)\textsuperscript{\hyperlink{tt19}{19}}        & 0 &  \xmark           & Direct      & -- &          &  23.9 & 23.2 & / --           &   -- \\
MPT-Instruct (7B)        & 5 &  \xmark           & CoT \#1      & -- &          &  28.8 & 31.7 & / --           &   -- \\
MPT-Instruct (30B)        & 0 &  \xmark           & Direct      & -- &          &  35.1  & 34.6 &/ --            &   -- \\
MPT-Instruct (30B)        & 5 &  \xmark           & CoT \#1      & -- &          &  40.1 & 40.3  &/ -- 
        &   -- \\
\addlinespace[0.15cm]
Falcon-Instruct (7B)\textsuperscript{\hyperlink{tt18}{18}}         & 0 &  \xmark           & Direct      & -- &         &  25.3  & 25.2  & / --           &   -- \\
Falcon-Instruct (7B)        & 5 &  \xmark           & CoT \#1      & -- &       & 24.0   &  23.8 & / --           &   -- \\
Falcon-Instruct (40B)        & 0 &  \xmark           & Direct      & -- &      & 39.0   &  30.0  & / --           &   -- \\
Falcon-Instruct (40B)        & 5 &  \xmark           & CoT \#1      & -- &       & 40.3  & 44.0 &/ -- 
        &   -- \\
\addlinespace[0.15cm]
Guanaco (33B)\textsuperscript{\hyperlink{tt17}{17}}         & 0 &  \xmark           & Direct      & -- &            &  42.9  & 37.4 &/ --            &   -- \\
Guanaco (33B)        & 5 &  \xmark           & CoT \#1      & -- &           &  48.0  & 40.3 &/ --            &   -- \\
Guanaco (65B)        & 0 &  \xmark           & Direct      & -- &           &  40.8 & 36.7 &/ --            &   -- \\
Guanaco (65B)        & 5 &  \xmark           & CoT \#1      & -- &           &  49.9 & 43.3 & / --           &   -- \\
\addlinespace[0.15cm]
Vicuna 1.3 (7B)\textsuperscript{\hyperlink{tt15}{15}}         & 0 &  \xmark           & Direct      & -- &            &  27.2 & 21.2 &/ --            &   -- \\
Vicuna 1.3 (7B)        & 5 &  \xmark           & CoT \#1      & -- &          &  39.7 & 33.6  &/ --            &   -- \\
Vicuna 1.3 (13B)        & 0 &  \xmark           & Direct      & -- &            &  38.7 & 38.3 &/ --            &   -- \\
Vicuna 1.3 (13B)        & 5 &  \xmark           & CoT \#1      & -- &          &  46.4 & 43.6  &/ --            &   -- \\
Vicuna 1.3 (33B)        & 0 &  \xmark           & Direct      & -- &            &  45.2 & 38.0 &/ --            &   -- \\
Vicuna 1.3 (33B)        & 5 &  \xmark           & CoT \#1      & -- &          &  49.2 & 41.3  &/ --            &   -- \\
Vicuna 1.3 (33B)        & 5 &  \xmark           & CoT \#1 + MV($k$=12)\textsuperscript{\hyperlink{tt10}{10}}      & -- &          &  52.2 & 44.7  &/ --            &   -- \\
\addlinespace[0.15cm]
Llama-2 (7B)\textsuperscript{\hyperlink{tt16}{16}}         & 0 &  \xmark           & Direct      & -- &          &  26.1 & 22.6  &/ --            &   -- \\
Llama-2 (7B)        & 5 &  \xmark           & CoT \#1      & -- &           &  34.1 & 36.2 & / --           &   -- \\
Llama-2 (7B)   & 5 &  \xmark           &  CoT \#1 + MV($k$=100)\textsuperscript{\hyperlink{tt10}{10}}     & -- &   
& 37.6      & 37.5 &/ --            &   -- \\
Llama-2 (13B)        & 0 &  \xmark           & Direct      & -- &           &  31.1 &  31.7 &/ --           &   -- \\
Llama-2 (13B)        & 5 &  \xmark           & CoT \#1      & -- &           &  40.0 & 42.8 &  / --          &   -- \\
Llama-2 (13B)   & 5 &  \xmark           &  CoT \#1 + MV($k$=100)\textsuperscript{\hyperlink{tt10}{10}}     & -- &   
& 46.7      & 45.5 &/ --            &   -- \\
Llama-2 (70B)        & 0 &  \xmark           & Direct      & -- &            &  43.4 & 42.8 & / --           &   -- \\
Llama-2 (70B)        & 5 &  \xmark           & CoT \#1      & -- &          &  57.4 & 53.6  & / --           &   -- \\
Llama-2 (70B)   & 5 &  \xmark           &  CoT \#1 + MV($k$=50)\textsuperscript{\hyperlink{tt10}{10}}     & -- &   
& $\star$62.5      & -- &/ --            &   -- \\
\addlinespace[0.15cm]
Llama-2-chat (7B)        & 0 &  \xmark           & Direct      & -- &          &  29.7 & 35.6  &/ --            &   -- \\
Llama-2-chat (7B)        & 5 &  \xmark           & CoT \#1      & -- &           &  32.9 & 33.2 & / --           &   -- \\
Llama-2-chat (13B)        & 0 &  \xmark           & Direct      & -- &           &  32.2 & 36.6 & / --           &   -- \\
Llama-2-chat (13B)        & 5 &  \xmark           & CoT \#1      & -- &           &  44.5 & 44.6 &  / --          &   -- \\
Llama-2-chat (70B)        & 0 &  \xmark           & Direct      & -- &            &  42.3 & 41.8 & / --           &   -- \\
Llama-2-chat (70B)        & 5 &  \xmark           & CoT \#1      & -- &          &  43.4 & 44.9  & / --           &   -- \\
\addlinespace[0.15cm]
Vicuna 1.5 (7B)\textsuperscript{\hyperlink{tt15}{15}}        & 0 &  \xmark           & Direct      & -- &          &  37.1 & 35.5  &/ --            &   -- \\
Vicuna 1.5 (7B)        & 5 &  \xmark           & CoT \#1      & -- &           &  40.5 & 41.2 &/ --            &   -- \\
Vicuna 1.5 (13B)        & 0 &  \xmark           & Direct      & -- &           &  41.6 & 41.5 &/ --            &   -- \\
Vicuna 1.5 (13B)        & 5 &  \xmark           & CoT \#1      & -- &           &  50.8 & 46.0 &/ --           &   -- \\
Vicuna 1.5 (13B)   & 0 &  \xmark           &  Direct  + MV($k$=100)\textsuperscript{\hyperlink{tt10}{10}}     & -- &   
& 41.7      & 42.6 &/ --            &   -- \\
Vicuna 1.5 (13B)   & 5 &  \xmark           &  CoT \#1 + MV($k$=100)\textsuperscript{\hyperlink{tt10}{10}}     & -- &   
& 50.4      & 46.3 &/ --            &   -- \\
\midrule
U-PaLM (540B)~\textsuperscript{\hyperlink{tt2}{2}}         & 5 &  \xmark           &  Direct  & 87.1 &/ --       & --        &  -- &/ --             & -- \\
U-PaLM (540B)~\textsuperscript{\hyperlink{tt2}{2}}         & 5 &  \xmark           &  CoT \#1     & 58.1 &/ --       & --        &  -- &/ --             & -- \\
Flan-U-PaLM (540B)~\textsuperscript{\hyperlink{tt2}{2}}    & 5 &  \xmark           &  Direct  & \underline{\textbf{90.3}} &/ --       & --        &  -- &/ --             & -- \\
Flan-U-PaLM (540B)~\textsuperscript{\hyperlink{tt2}{2}}    & 5 &  \xmark           &  CoT \#1     & 80.6 &/ --       & --        &  -- &/ --             & -- \\

Med-PaLM V2 (540B)~\textsuperscript{\hyperlink{tt22}{22}}    & \textit{finetuned} &  \xmark           &  --     & -- &       & \underline{\textbf{86.5}}         &  72.3 &/ --             & \underline{\textbf{81.8}} \\

\midrule
PubMedBERT (110M)~\textsuperscript{\hyperlink{t1}{1}}   & \textit{finetuned}      &  \xmark           & --           & -- &/ --        & --        & 40.0 &/ 41.0          & -- \\
PubMedBERT (110M)~\textsuperscript{\hyperlink{tt3}{3}}   & \textit{finetuned}     &  \cmark           & --            & -- &/ --       & --        & 43.0 &/ 47.0          & -- \\
BioLinkBERT (345M)~\textsuperscript{\hyperlink{tt4}{4}}  & \textit{finetuned} &  \cmark           & --            & -- &/ 50.7       & 44.6        & -- &/ --              &  72.2 \\
\addlinespace[0.15cm]
BioGPT (347M)\textsuperscript{\hyperlink{tt11}{11}} & \textit{finetuned} & \xmark & -- & -- &/-- & -- & -- &/-- & 78.2 \\
PubMedGPT (2.7B)\textsuperscript{\hyperlink{tt12}{12}} & \textit{finetuned} & \xmark & -- & -- &/-- & 50.3 & -- &/-- & 74.4 \\
Galactica (120B)\textsuperscript{\hyperlink{tt13}{13}} & \textit{finetuned} &  \xmark &  -- & -- &/-- & 44.4 & 52.9 &/ -- & 77.6 \\
\midrule
Human (passing score)  & --                         & --                & --            & \multicolumn{2}{c}{60.0~\textsuperscript{\hyperlink{tt5}{5}}}  & 60.0~\textsuperscript{\hyperlink{tt5}{5}}   & \multicolumn{2}{c}{50.0~\textsuperscript{\hyperlink{tt7}{7},\hyperlink{tt8}{8}}} & -- \\
Human (expert score)  & --                                &  --               & --            & \multicolumn{2}{c}{87.0~\textsuperscript{\hyperlink{tt6}{6}}}  & 87.0~\textsuperscript{\hyperlink{tt6}{6}} & \multicolumn{2}{c}{90.0~\textsuperscript{\hyperlink{tt3}{3}}} & 78.0~\textsuperscript{\hyperlink{tt9}{9}} \\
\bottomrule
\addlinespace[0.15cm]
\multicolumn{10}{l}{\small{
\hypertarget{tt1}{\textsuperscript{1}}~\textsl{professional medicine subset}~(USMLE, \cite{Hendrycks2020-jw}),
~\hypertarget{tt2}{\textsuperscript{2}}~\cite{Chung2022-lf},
~\hypertarget{tt3}{\textsuperscript{3}}~\cite{Pal2022-ph},
~\hypertarget{tt4}{\textsuperscript{4}}~\cite{Yasunaga2022-gl}
}}\\
\multicolumn{10}{l}{\small{
\hypertarget{tt5}{\textsuperscript{5}}~USMLE (passing score): \url{https://www.usmle.org/scores-transcripts}
,~\hypertarget{tt6}{\textsuperscript{6}}~USMLE (expert score): 95th percentile~\citep{Hendrycks2020-jw}
}}\\
\multicolumn{10}{l}{\small{
\hypertarget{tt7}{\textsuperscript{7}}~MedMCQA test (AIIMS): \url{https://collegedunia.com/exams/aiims-mbbs/cutoff}
}}\\
\multicolumn{10}{l}{\small{
\hypertarget{tt8}{\textsuperscript{8}}~MedMCQA valid. (NEET PG): \url{https://medicine.careers360.com/articles/neet-pg-cut-off}
}}\\
\multicolumn{10}{l}{\small{
\hypertarget{tt9}{\textsuperscript{9}}~\cite{Jin2019-qa},~\hypertarget{tt10}{\textsuperscript{10}}~Majority voting classifier with $k$ samples and temperatur $\tau=0.5$ for Codex, $\tau=0.9$ for all other models (self-consistency~\cite{Wang2022-jx})
}} \\
\multicolumn{10}{l}{\small{
\hypertarget{tt11}{\textsuperscript{11}}~\cite{Luo2022-qh}
,~\hypertarget{tt12}{\textsuperscript{12}}~\cite{Venigalla2022-dd}
,~\hypertarget{tt13}{\textsuperscript{13}}~\cite{Taylor2022-ws}
,~\hypertarget{tt14}{\textsuperscript{14}}~\cite{Robinson2022-uh}
,~\hypertarget{tt15}{\textsuperscript{15}}~\cite{zheng2023judging}
,~\hypertarget{tt16}{\textsuperscript{16}}~\cite{touvron2023llama}
,~\hypertarget{tt17}{\textsuperscript{17}}~\cite{dettmers2023qlora}
}}\\
\multicolumn{10}{l}{\small{
\hypertarget{tt17}{\textsuperscript{18}}~\cite{falcon40b}
,~\hypertarget{tt119}{\textsuperscript{19}}~\cite{MosaicML2023Introducing}
,~\hypertarget{tt120}{\textsuperscript{20}}~\cite{GPT-NeoX-20B}
,~\hypertarget{tt121}{\textsuperscript{21}}~\cite{Nori2023CapabilitiesOG}
,~\hypertarget{tt122}{\textsuperscript{22}}~\cite{singhal2023expertlevel}
}}
\end{tabular}
}
\end{center}
\end{table}

\clearpage
\section{Datasets}\label{apdx:datasets}
\paragraph{MedQA-USMLE} 
\cite{Jin2021-jo} gathers historical questions from the United States Medical Licensing Examination (USMLE), which targets trained medical professionals. The questions are notorious for being challenging as they often require strong problem-solving skills coupled with comprehensive medical knowledge. Each question features a description of a medical case and a question that emulates the real clinical setting. The more recent MMLU dataset~\citep{Hendrycks2020-jw} has 31 validation and 272 test USMLE questions (around 105 words/question). In Appendix \ref{apdx:medqa-vs-mmlu}, we benchmark both USMLE datasets and found the MedQA USMLE dataset to be more difficult. The MedQA-USMLE data does not come with explanations. Instead, we use the MMLU-USMLE CoTs from \cite{Chung2022-lf} that are available from {\url{https://github.com/jasonwei20/flan-2}}.

 \paragraph{MedMCQA} 
 \cite{Pal2022-ph} is a large-scale multiple-choice question answering collected from Indian medical school entrance exams (AIIMS and NEET-PG). The MedMCQA covers a broad range of medical topics (dentistry, psychiatry, surgery, \ldots) and requires being able to follow a variety of reasoning types (logic, factual, comparison, \ldots). However, questions are often more knowledge-centred than the USMLE questions, which tend to focus more on problem-solving skills.
 
 \paragraph{PubMedQA} 
\cite{Jin2019-qa} is a collection of expert-annotated yes/no/maybe research questions derived from PubMed abstracts. Whereas the questions from the USMLE and the MedMCQA datasets are self-contained and might be answered using general medical knowledge and methodology, each PubMedQA question is contextualized on a provided abstract. Therefore PubMedQA primarily focuses on evaluating reading comprehension skills.

\section{MedQA-USMLE versus MMLU-USMLE}\label{apdx:medqa-vs-mmlu}

\begin{multicols}{2}
\begin{table}[H]
\caption{Comparing the USMLE datasets from (test) MedQA~\citep{Jin2021-jo} and (validation/test) MMLU~\citep{Hendrycks2020-jw}. We include the results of the recent Flan-U-PaLM 540B~\citep{Chung2022-lf}. All models use 5 shots.}
\label{tab:medqa-vs-mmlu}
\vspace{-1.0em}
\begin{center}
\resizebox{\columnwidth}{!}{%
\begin{tabular}{l@{\hspace{-16pt}}lr@{\hspace{1pt}}l c}
\toprule
\bf Model & \bf Prompt & \multicolumn{2}{c}{\textbf{MMLU}} & \bf MedQA \\
\midrule
Codex  &  Direct  & 77.2 &/ \underline{\textbf{70.1}}   & \underline{\textbf{56.6}}     \\ 
Codex  &  CoT \#1     & 80.6 &/ 69.1   & 56.2      \\ 
U-PaLM       &  Direct  & 87.1 &/ --       & --        \\
U-PaLM         &  CoT \#1     & 58.1 &/ --       & --         \\
Flan-U-PaLM    &  Direct  & \underline{\textbf{90.3}} &/ --       & --     \\
Flan-U-PaLM    &  CoT \#1     & 80.6 &/ --       & --    \\
\midrule
Human (passing score)  &             & \multicolumn{2}{c}{60.0}  & 60.0 \\
Human (expert score)   &             & \multicolumn{2}{c}{87.0}  & 87.0  \\
\bottomrule
\end{tabular}
}
\end{center}
\end{table}

In Table \ref{tab:medqa-vs-mmlu}, we report the performances of the three medical question answering datasets as well as the \textit{professional medicine} subset of the MMLU dataset~\citep{Hendrycks2020-jw}, which was also explored in recent related work~\citep{Chung2022-lf}.

Based on Codex performances, the MedQA-USMLE dataset appears to be more challenging than the MMLU-USMLE counterpart. Codex~\citep{Chen2021-sj} in a 5-shot setting (Direct and CoT prompting, $\tau$=0), scores around 13.2\% lower accuracy on the MedQA-USMLE ($\sim$56.4\%) than on the MMLU-USMLE  ($\sim$69.6\%). Succeeding the USMLE requires a score of around 60\%.

\end{multicols}

\clearpage
\section{Comparing GPT versions on the USMLE dataset}\label{apdx:gpt-versions-comparison}

We report the test USMLE accuracy for multiple GPT version in Table \ref{tab:usmle-results-smaller-gpt} for the direct and CoT \#1 prompts. Note that Codex (\texttt{code-davinci-002}) is a large GPT-3 model pre-trained on text and code; InstructGPT (\texttt{text-davinci-002}) is a version of Codex finetuned based human-feedback to ``follow the user’s instructions helpfully and safely''.\footnote{\url{https://beta.openai.com/docs/model-index-for-researchers}}

\begin{multicols}{2}
\vspace{0.5em}
\begin{table}[H]
\caption{Answering accuracy of multiple GPT-3 models on the USMLE dataset in a zero-shot setting.}
\label{tab:usmle-results-smaller-gpt}
\begin{center}
\vspace{-1.0em}
\resizebox{\columnwidth}{!}{%
\begin{tabular}{llcc}
\toprule
\bf Model & \bf Prompt & \bf Acc. & $\Delta$  \\
\midrule
\texttt{text-curie-001}  & Direct &    27.8 & {\color{nicered} -9.4 } \\
\texttt{text-davinci-001}  & Direct &    37.2 &  -- \\
\texttt{code-davinci-002}  & Direct &   52.5 &  {\color{niceblue} +15.3}\\
\texttt{text-davinci-002}  & Direct &   46.0 &   {\color{niceblue} +8.8 } \\
\midrule
\texttt{text-curie-001}  & CoT \#1  &    25.5 & {\color{nicered} -14.7 } \\ 
\texttt{text-davinci-001}  & CoT \#1  &    40.2 & -- \\
\texttt{code-davinci-002}  & CoT \#1  &    52.9 & {\color{niceblue} +12.7}  \\
\texttt{text-davinci-002}  & CoT \#1  &    47.1 & {\color{niceblue} +6.9 }\\
\midrule
Random & & 25.0 & \\
\bottomrule
\end{tabular}
}
\end{center}
\end{table}
\vspace{-1.0em}
The smallest model performed only slightly better than at random, with an accuracy of maximum 27.8\% for the \texttt{curie} model, whereas the largest model non-aligned text model  \texttt{text-davinci-001} scored a maximum of 40.2\% for all prompts. The best performances are obtained with the text and code pre-trained model \texttt{code-davinici-002} (52.9\%). Human-alignment appears to damage answering performances: \texttt{text-davinci-002} scored a maximum of 47.1\%.  This suggests that advanced medical reasoning capabilities only emerge in the largest of the GPT-3 models, and that code pre-trained is highly effective. In this experiment, human-alignment led to a decrease of accuracy, although we found InstructGPT to overall produce more readable samples than Codex in zero-shot CoT settings (Appendix \ref{apdx:additional-CoT-samples}).
\end{multicols}

\section{Answering bias}\label{apdx:answering-bias}

\begin{table}[H]
\caption{Frequencies of predictions and labels. Classification bias of InstructGPT and Codex on the USMLE dataset, with (\cmark) and without (\xmark) random label permutation. We highlight labels that are under estimated using the color {\color{niceblue} blue $\blacktriangledown$} and over estimated using the color {\color{nicered} red $\blacktriangleup$} ($\pm$ 10\% of the label frequency). Using the $\chi^2$ test, we report the pp-value for the null hypothesis "\textit{the predictive distribution equals the empirical one}". The models are evaluated using zero shot and $T=0$, unless specified.}
\label{tab:usmle-bias}
\begin{center}
\vspace{0.0em}
\resizebox{0.8\columnwidth}{!}{%
\begin{tabular}{cllccccll}
\toprule
\bf Perm. & \bf Model & \bf Prompt & \bf A & \bf B & \bf C & \bf D & \bf Acc. & \bf $p$-value \\
\midrule
\xmark & InstructGPT & Direct &  {\color{niceblue} 155$\blacktriangledown$}  & 299 & {\color{nicered} 405$\blacktriangleup$}  &  {\color{nicered} 414$\blacktriangleup$} & 46.0 & $< 10 ^{-10}$ \\
\xmark  & InstructGPT & CoT \#1  & {\color{nicered}421$\blacktriangleup$} & {\color{niceblue}240$\blacktriangledown$}  & {\color{niceblue}291$\blacktriangledown$}  & {\color{nicered}321$\blacktriangleup$} & 47.1 & $1 \cdot 10 ^{-10}$ \\
\xmark  & InstructGPT & CoT \#2  & {\color{nicered}423$\blacktriangleup$} & {\color{niceblue}211$\blacktriangledown$}  & {\color{niceblue}286$\blacktriangledown$}  & {\color{nicered}353$\blacktriangleup$} & 46.8 & $< 10 ^{-10}$ \\
\xmark  & InstructGPT & CoT \#3  & {\color{nicered}416$\blacktriangleup$} & {\color{niceblue}236$\blacktriangledown$}  & {\color{niceblue}272$\blacktriangledown$} & {\color{nicered}349$\blacktriangleup$} & 46.0 & $< 10 ^{-10}$ \\
\xmark  & InstructGPT & CoT \#4  & 378  & {\color{niceblue}221$\blacktriangledown$}  & {\color{niceblue}294$\blacktriangledown$} & {\color{nicered}380$\blacktriangleup$}  & 45.6 & $< 10 ^{-10}$ \\
\xmark  & InstructGPT & CoT \#5  & {\color{nicered}392$\blacktriangleup$} & {\color{niceblue}234$\blacktriangledown$} & {\color{niceblue}277$\blacktriangledown$} & {\color{nicered}370$\blacktriangleup$} & 45.1 & $< 10 ^{-10}$ \\
\xmark  &  & \bf data & \bf 353 &\bf  309 & \bf 346 & \bf 265 & &  \\
\midrule
\cmark  & InstructGPT & Direct & {\color{niceblue}138$\blacktriangledown$}  & 295 & {\color{nicered}377$\blacktriangleup$}  & {\color{nicered}463$\blacktriangleup$}  & 46.5 & $< 10 ^{-10}$ \\
\cmark  & InstructGPT & CoT \#1  & {\color{nicered}374$\blacktriangleup$}  & {\color{niceblue}276$\blacktriangledown$}  &  {\color{niceblue}252$\blacktriangledown$} & {\color{nicered}371$\blacktriangleup$} & 45.3 & $4 \cdot 10 ^{-10}$ \\
\cmark  &  &  \bf data & \bf 317 & \bf 326 & \bf 323 & \bf 307 & \\
\midrule
\xmark  & Codex & Direct                        & {\color{niceblue}163$\blacktriangledown$} & {\color{nicered}360$\blacktriangleup$} & {\color{nicered}407$\blacktriangleup$} & {\color{nicered}343$\blacktriangleup$} & 52.1 & $< 10 ^{-10}$ \\
\xmark  & Codex (5 shots)  & Direct             & {\color{niceblue}254$\blacktriangledown$} & {\color{niceblue}285$\blacktriangledown$} & {\color{nicered}430$\blacktriangleup$} & {\color{nicered}304$\blacktriangleup$} & 56.6 & $< 10 ^{-10}$ \\
\xmark  & Codex & CoT \#1                       & 315 & {\color{niceblue}250$\blacktriangledown$} & 285 & {\color{nicered}423$\blacktriangleup$} & 52.9 & $< 10 ^{-10}$ \\
\xmark  & Codex  (5 shots) & CoT \#1            & 334 & 300 & 324 & {\color{nicered}315$\blacktriangleup$} & 56.2 & $7 \cdot 10 ^{-03}$ \\
\xmark  & Codex (5 shots, $\tau$=0.5)\textsuperscript{1}  & CoT \#1   & 340 & 281 & {\color{niceblue}308$\blacktriangledown$}  & {\color{nicered}349$\blacktriangleup$} & 60.2 & $2 \cdot 10 ^{-06}$ \\
\xmark  &  & \bf data & \bf 353 &\bf  309 & \bf 346 & \bf 265 & \\
\bottomrule
\addlinespace[0.1cm]
\multicolumn{8}{l}{\small{
\textsuperscript{1}Averaged using $k=100$ samples.
}}
\end{tabular}
}
\end{center}
\end{table}

\begin{figure}[h]
    \centering
    \caption{Frequencies of predicted labels for Codex 5-shot CoT (average of $k$=100 samples) and ground truth label frequencies. For the USMLE, we report frequencies of the zero-shot InstructGPT (Direct and CoT prompting), originally displayed in Figure \ref{fig:usmle-gpt3-bias}.}
    \label{fig:codex-bias}
    \resizebox{1.0\columnwidth}{!}{%
        \includegraphics{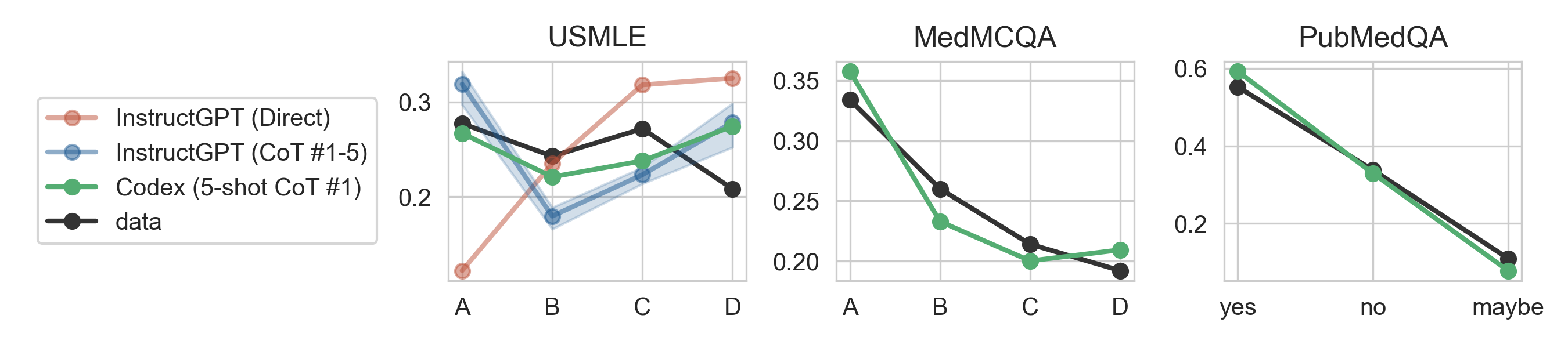}
    }
\end{figure}

In Table \ref{tab:usmle-bias}, we report the frequencies of answers and of predicted labels with and without label permutation. We report the frequencies for InstructGPT as well as Codex.

Querying InstructGPT using the CoT prompts resulted in a more faithful predictive distribution of the labels. Nonetheless, a bias towards the labels A and D and a tendency to avoid predicting labels B and C could still be observed. To confirm whether this bias originates from the data or the model, we permuted the labels and repeated the experiment for prompts number 0 and 1 and observed the same trend. Codex exhibits similar trends, although few-shot learning seems to yield more faithful predictive distributions.

In all cases, models tend to default to the label D. In Figure \ref{fig:usmle-sample-3}, we present two CoT leading to mispredicted label D. In both cases, GPT-3 fails to narrow down to one answer options and defaults to option D.

Figure \ref{fig:codex-bias} presents some of the results from Table \ref{tab:usmle-bias} for the USMLE and extend it with the frequencies observed in the two other datasets for Codex 5-shot CoT ($k=100$ samples). The bias appears less important for the MedMCQA and PubMedQA datasets than for the USMLE dataset.

\section{Information retrieval}\label{apdx:information-retrieval}

Wikipedia articles were converted into overlapping passages of size 100 words and indexed along with their respective article titles. Given a question $\rvq$, an answer choice $\rva$, and weights $\beta_1=1, \beta_2=1, \beta_3=0.5$. The weights were chosen based on a qualitative assessment of the retrieved passages on a few questions. we retrieved passages $\rvd$ based on a composite \bmtf score defined as
\begin{equation}
    \mathrm{score}(\rvq, \rva, \rvd) = \beta_1 \cdot  \mathrm{\bmtf}(\rvq, \rvd_{\mathrm{content}}) + \beta_2 \cdot \mathrm{\bmtf}(\rva,  \rvd_{\mathrm{content}}) + \beta_3 \cdot  \mathrm{\bmtf}(\rva, \rvd_{\mathrm{title}})  \ .
\end{equation}

\section{Open-Source LLMs}\label{apdx:open-source}

\begin{figure}[h]
    \centering
    \caption{Benchmarking LLMs on the MedQA-USMLE and MedMCQA datasets using direct 0-shot and 5-shot CoT prompting. All results are obtained using greedy decoding ($\tau=0$).}
    \label{fig:open-source-benchmark-all}
    \resizebox{\columnwidth}{!}{%
        \includegraphics{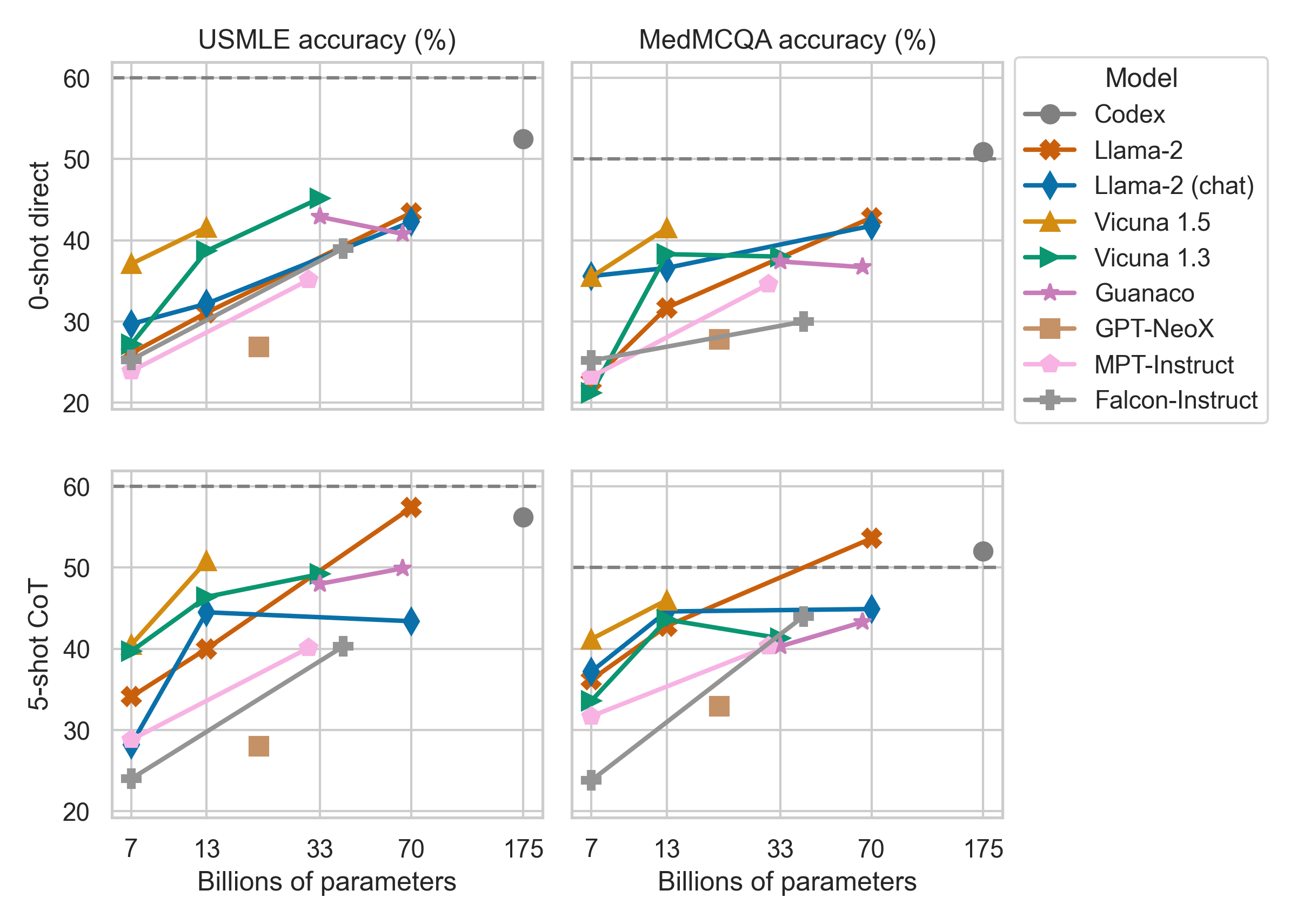}
    }
\end{figure}

We assessed the performances of open-source models (Vicuna, Guanaco, GPT-NeoX, MPT-instruct, Falcon and Llama-2) on the MedQA-USMLE and MedMCQA datasets in zero-shot and 5-shot settings, all using greedy decoding ($\tau = 0$). We report results in Figure \ref{fig:open-source-benchmark-all}.

\section{Additional CoT Samples}\label{apdx:additional-CoT-samples}

\paragraph{CoT prompt variations} 

\begin{figure}[H]
    \centering
    \caption{A selection of remarkable CoTs generated by InstructGPT \texttt{text-davinci-002} based on USMLE questions. GPT-3 can adopt diverse problem-solving strategies.}
    \label{fig:usmle-remarkable-strategies}
    \includegraphics[width=\columnwidth]{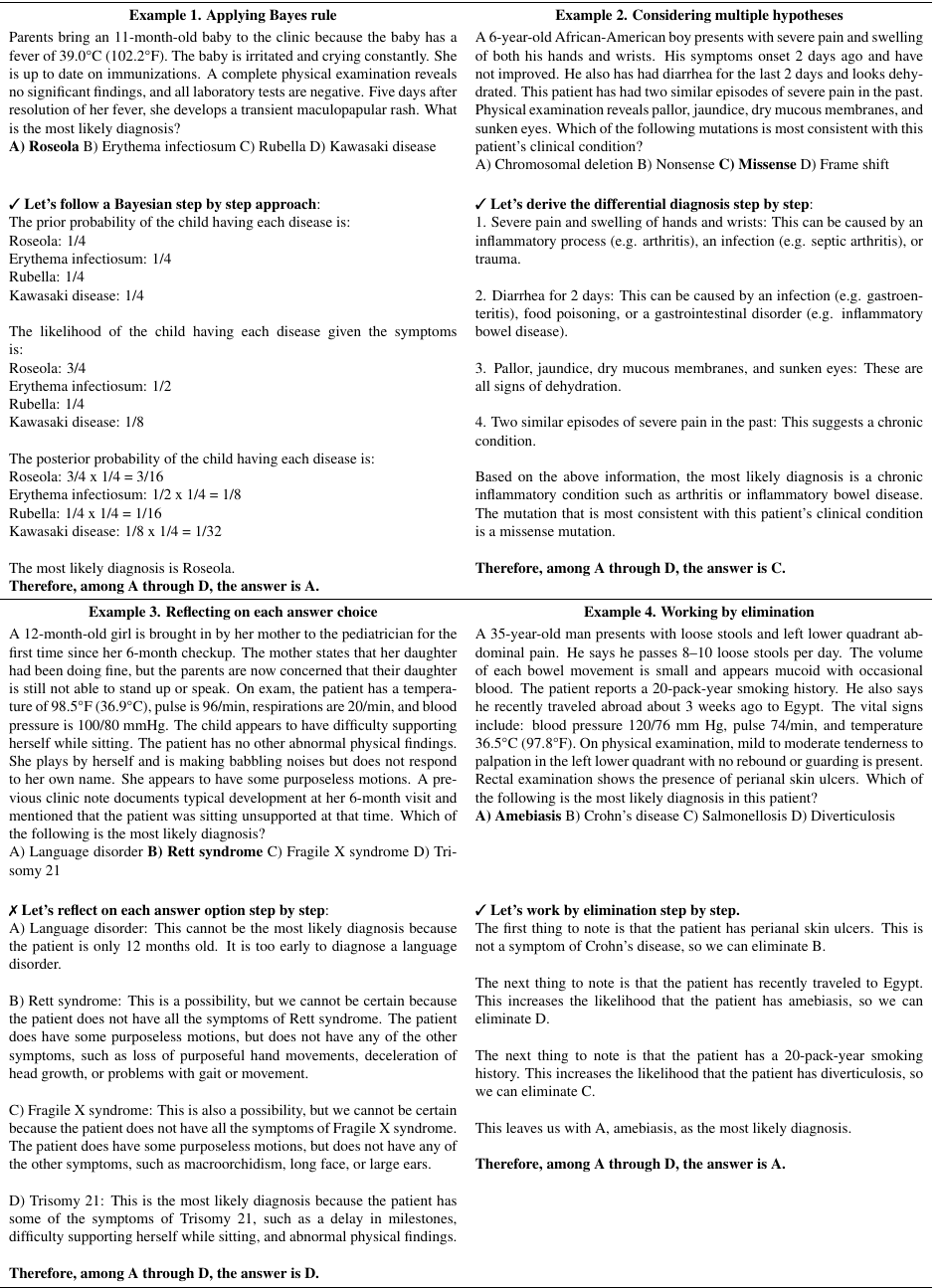}
\end{figure}

In Figure \ref{fig:usmle-remarkable-strategies}, we report four selected CoTs generated from the prompt variations studied in Appendix \ref{apdx:prompt-selection}

\paragraph{Codex CoTs}

\begin{figure}[H]
    \centering
    \caption{Two randomly selected CoTs generated by Codex \texttt{code-davinci-002} based on USMLE questions. Codex tends to repeat itself and generate zero-shot CoTs of lower quality than InstructGPT.}
\label{fig:usmle-codex-CoTs}
    \includegraphics[width=\columnwidth]{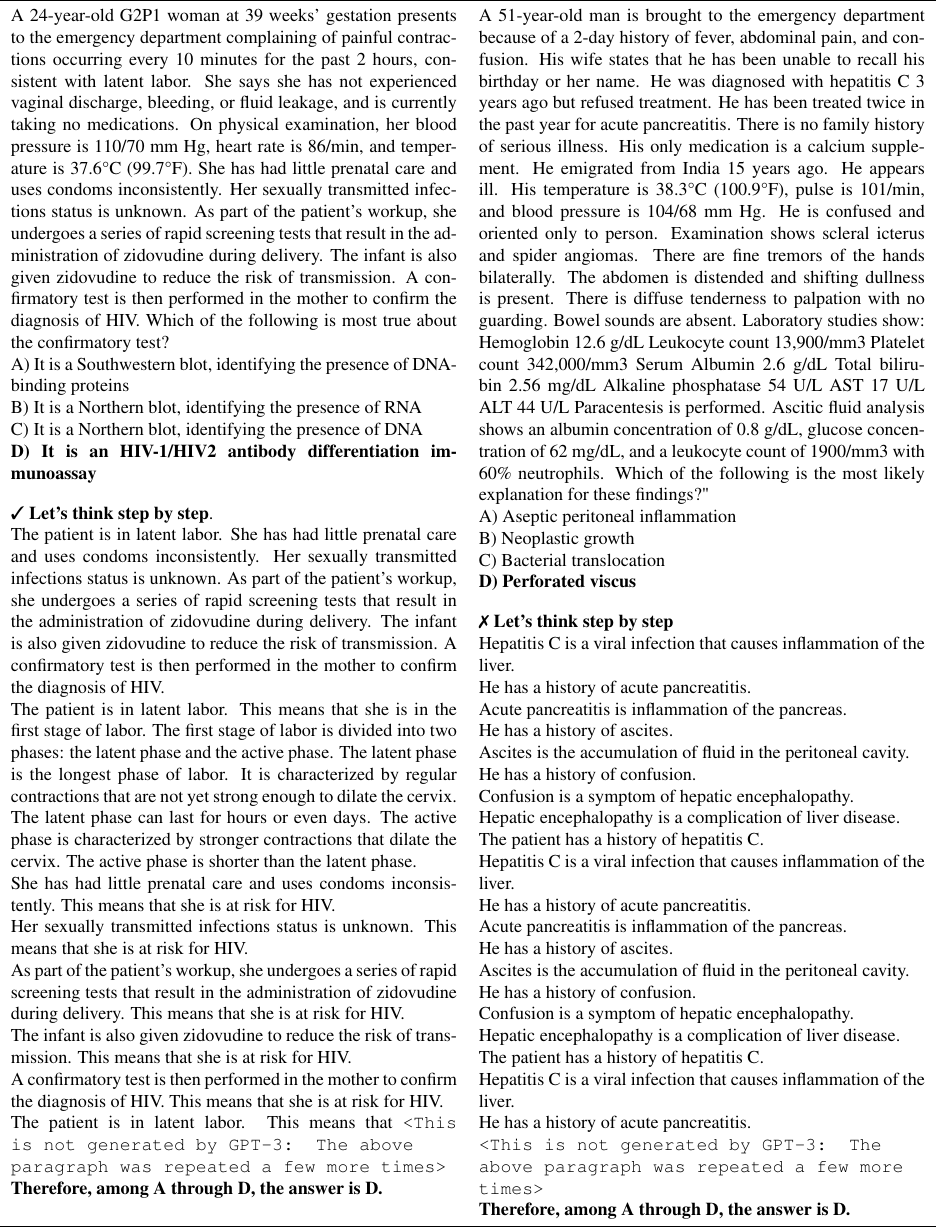}
\end{figure}

In Figure \ref{fig:usmle-codex-CoTs}, we display CoTs generated by Codex. Codex appears to yield CoTs of lower quality than InstructGPT (frequent repetitions, less verbosity).

\paragraph{Annotated InstructGPT CoTs} 

\begin{figure}[H]
    \centering
    \caption{(Sample 2) Generated zero-shot Chain-of-Thought from InstructGPT \texttt{text-davinci-002} for three CoT prompts on a sample for the MedQA-USMLE test set.}
    \label{fig:usmle-sample-2}
    \includegraphics[width=\columnwidth]{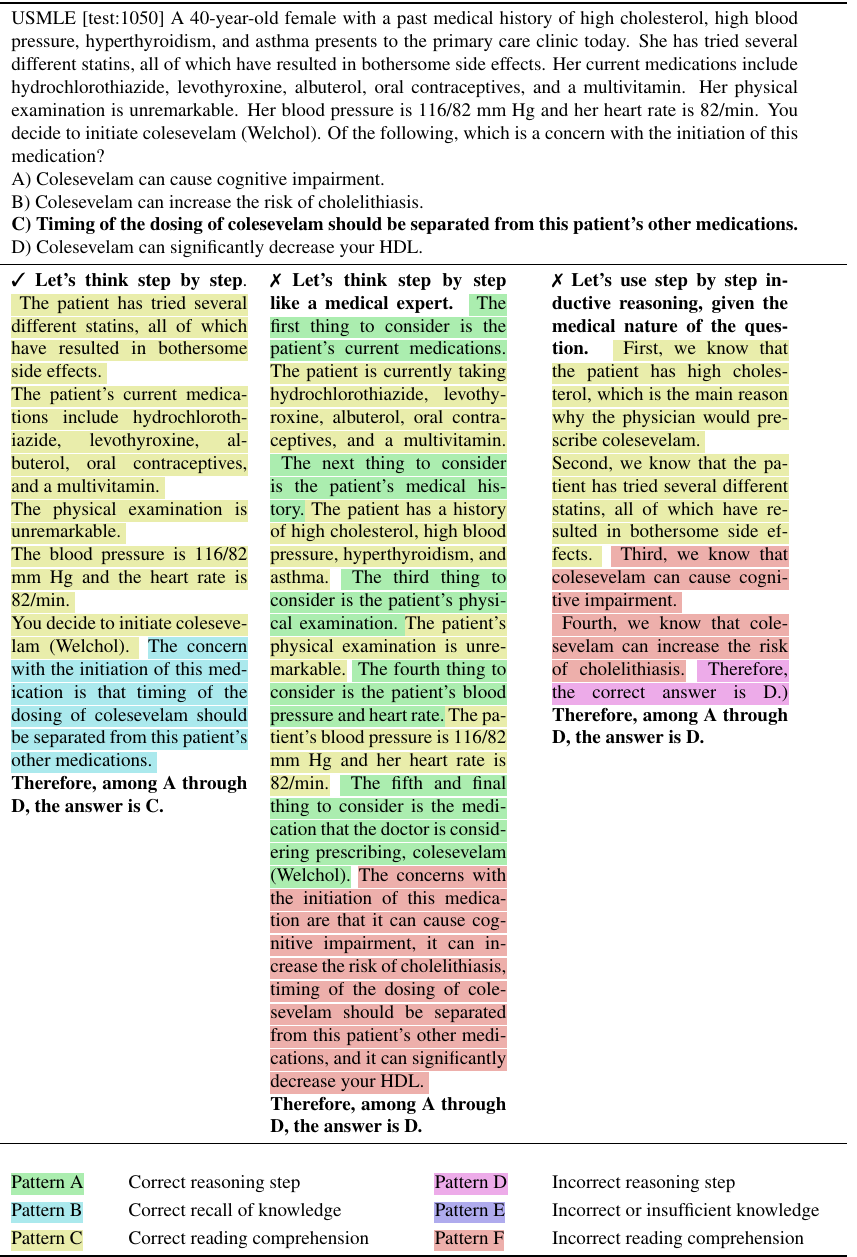}
\end{figure}

\begin{figure}[H]
    \centering
    \caption{(Sample 3) Generated zero-shot Chain-of-Thought from InstructGPT \texttt{text-davinci-002} for three CoT prompts on a sample for the MedQA-USMLE test set.}
    \label{fig:usmle-sample-3}
    \includegraphics[width=\columnwidth]{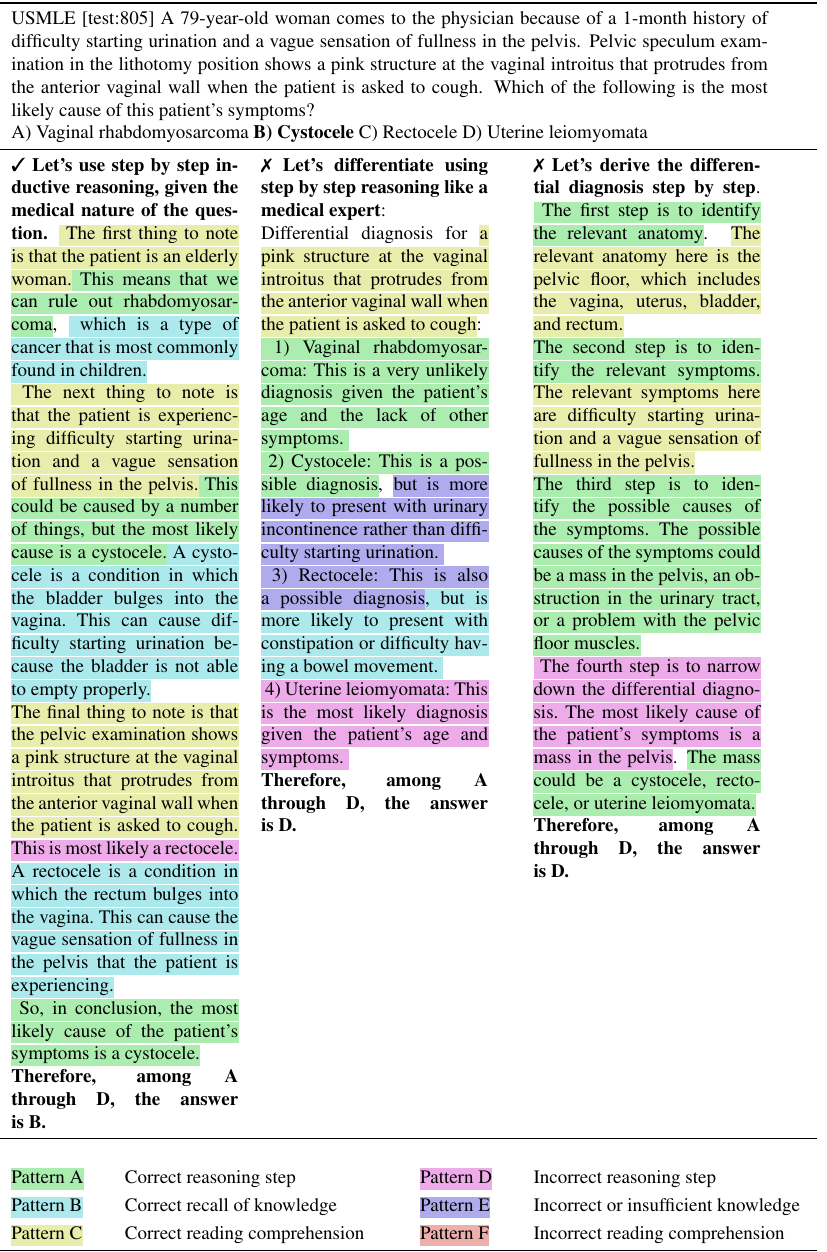}
\end{figure}

\begin{figure}[H]
    \centering
    \caption{(Sample 4) Generated zero-shot Chain-of-Thought from InstructGPT \texttt{text-davinci-002} for three CoT prompts on a sample for the MedQA-USMLE test set.}
    \label{fig:usmle-sample-4}
    \includegraphics[width=\columnwidth]{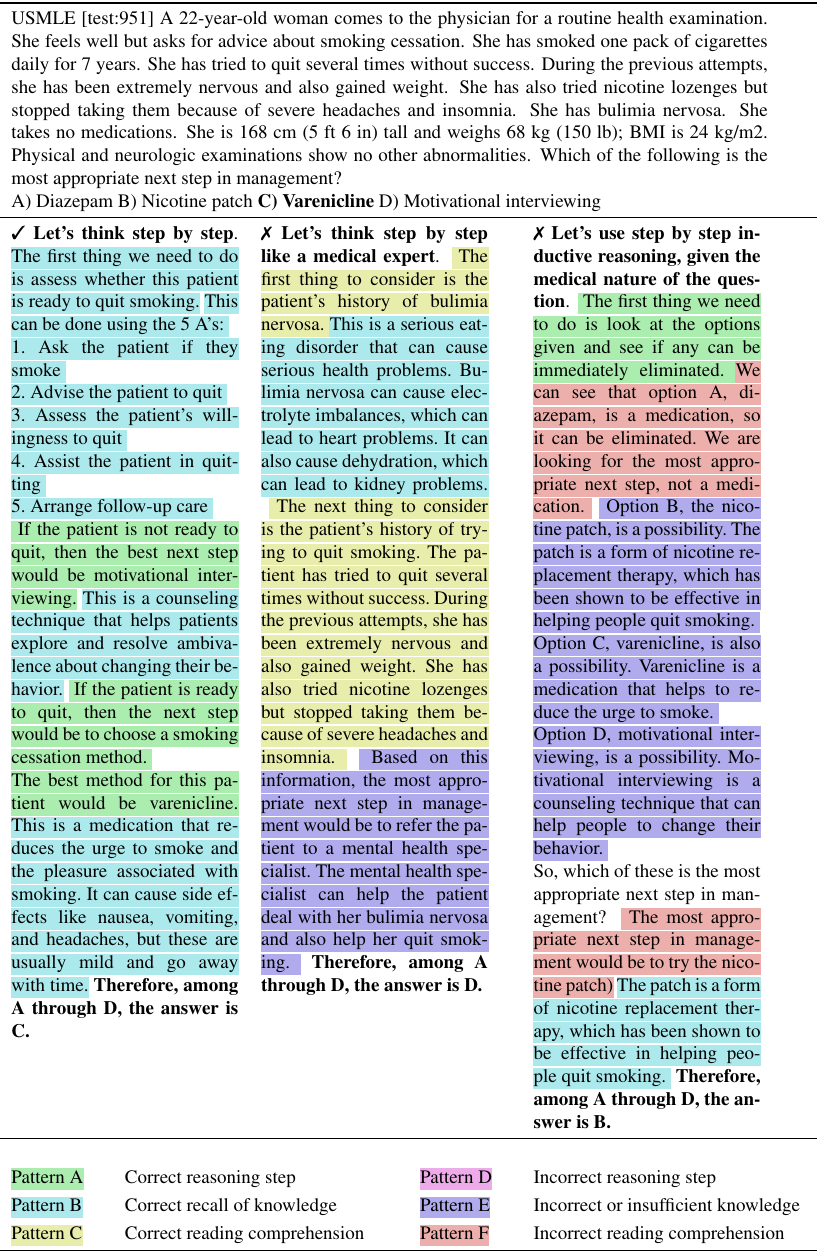}
\end{figure}

\begin{figure}[H]
    \centering
    \caption{(Sample 5) Generated zero-shot Chain-of-Thought from InstructGPT \texttt{text-davinci-002} for three CoT prompts on a sample for the MedQA-USMLE test set.}
    \label{fig:usmle-sample-5}
    \includegraphics[width=\columnwidth]{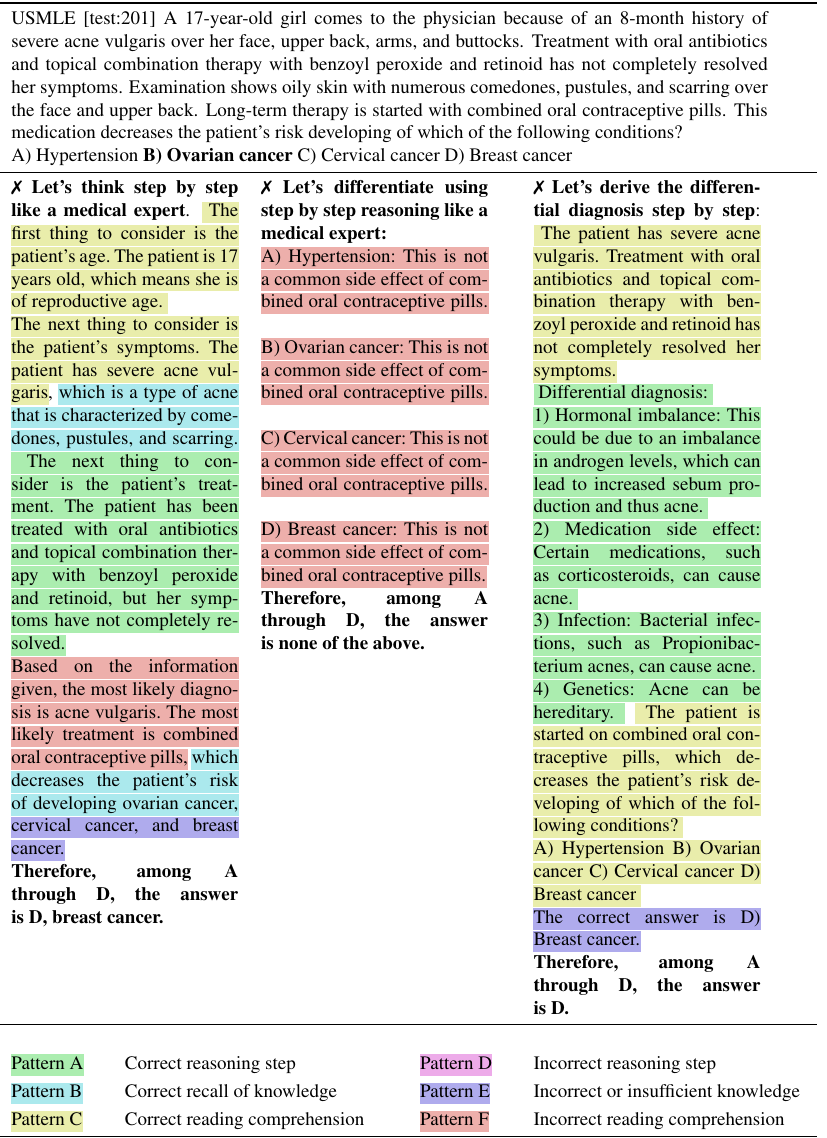}
\end{figure}

\begin{figure}[H]
    \centering
    \caption{(Sample 6) Generated zero-shot Chain-of-Thought from InstructGPT \texttt{text-davinci-002} for three CoT prompts on a sample for the MedQA-USMLE test set.}
    \label{fig:usmle-sample-6}
    \includegraphics[width=\columnwidth]{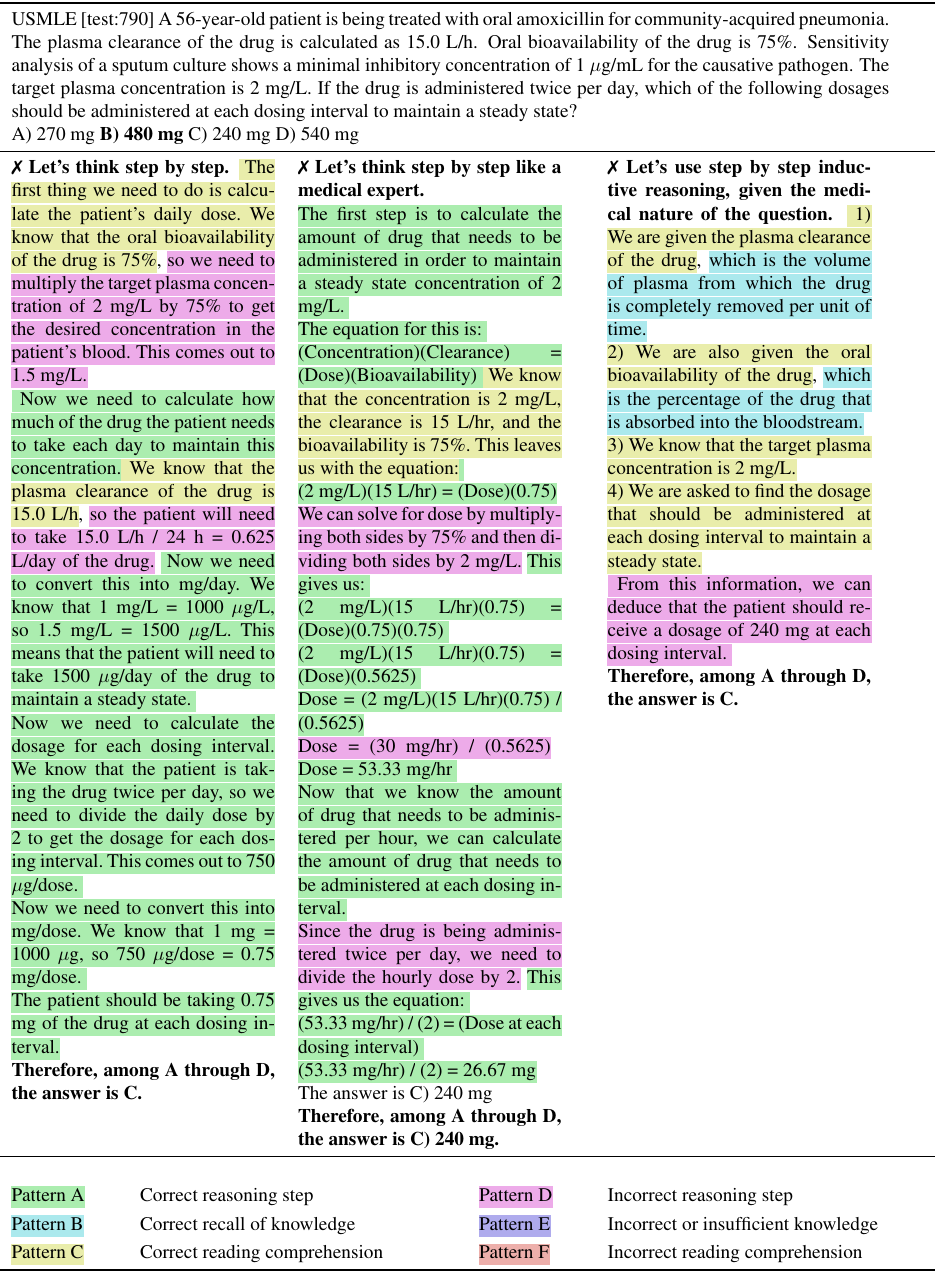}
\end{figure}

\begin{figure}[H]
    \centering
    \caption{(Sample 7) Generated zero-shot Chain-of-Thought from InstructGPT \texttt{text-davinci-002} for three CoT prompts on a sample for the MedQA-USMLE test set.}
    \label{fig:usmle-sample-7}
    \includegraphics[width=\columnwidth]{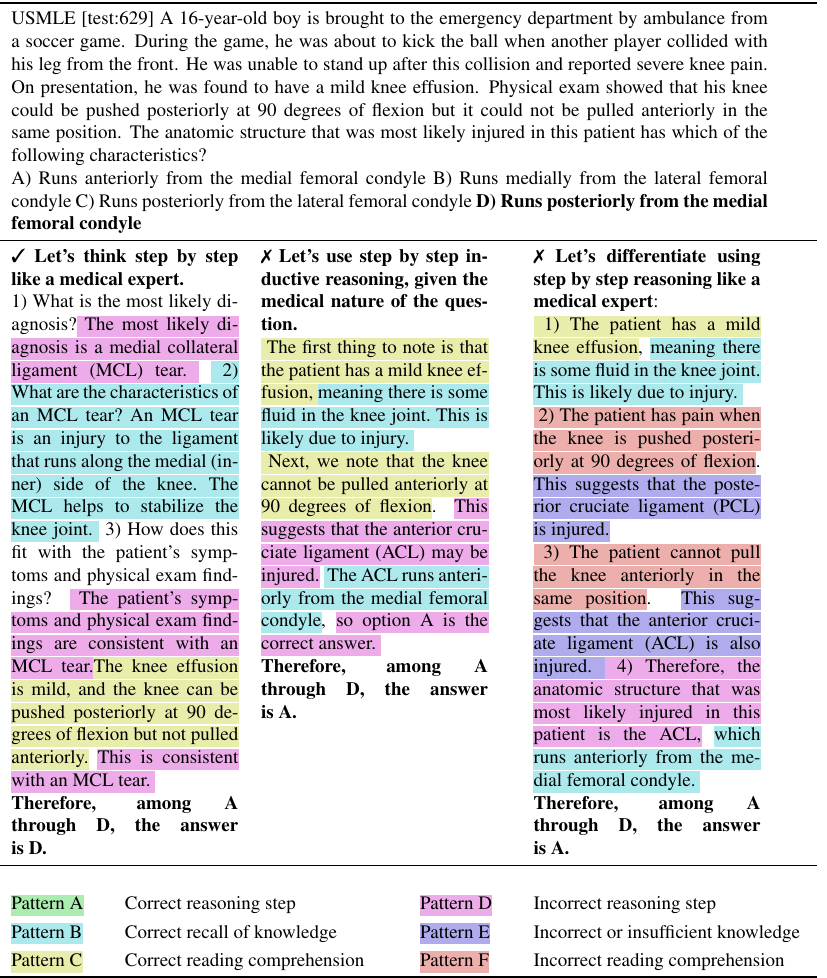}
\end{figure}

\begin{figure}[H]
    \centering
    \caption{(Sample 8) Generated zero-shot Chain-of-Thought from InstructGPT \texttt{text-davinci-002} for three CoT prompts on a sample for the MedQA-USMLE test set.}
    \label{fig:usmle-sample-8}
    \includegraphics[width=\columnwidth]{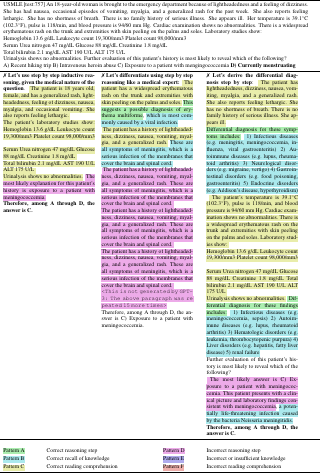}
\end{figure}

\begin{figure}[H]
    \centering
    \caption{(Sample 9) Generated zero-shot Chain-of-Thought from InstructGPT \texttt{text-davinci-002} for three CoT prompts on a sample for the MedQA-USMLE test set.}
    \label{fig:usmle-sample-9}
    \includegraphics[width=\columnwidth]{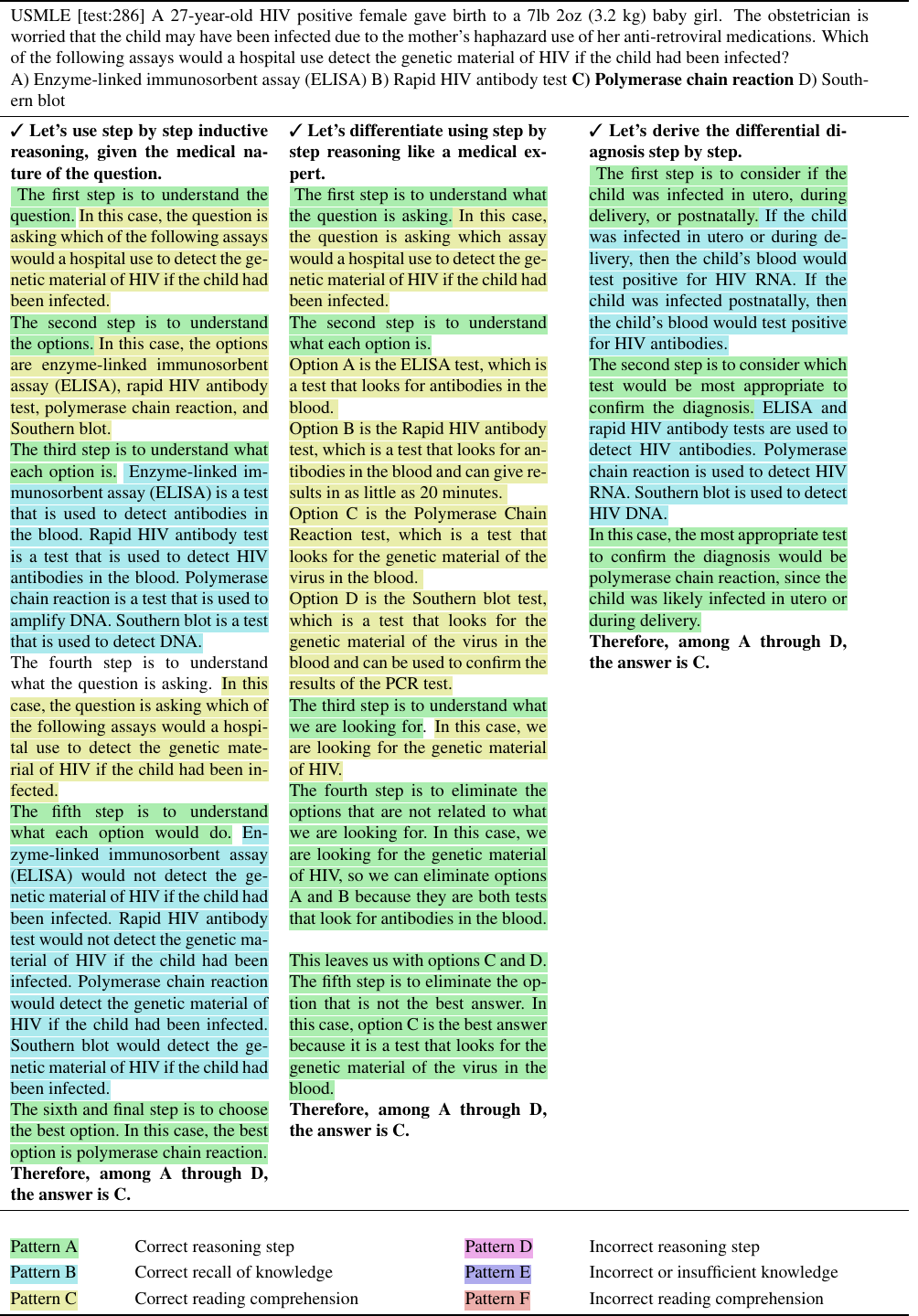}
\end{figure}

\begin{figure}[H]
    \centering
    \caption{(Sample 10) Generated zero-shot Chain-of-Thought from InstructGPT \texttt{text-davinci-002} for three CoT prompts on a sample for the MedQA-USMLE test set.}
    \label{fig:usmle-sample-10}
    \includegraphics[width=\columnwidth]{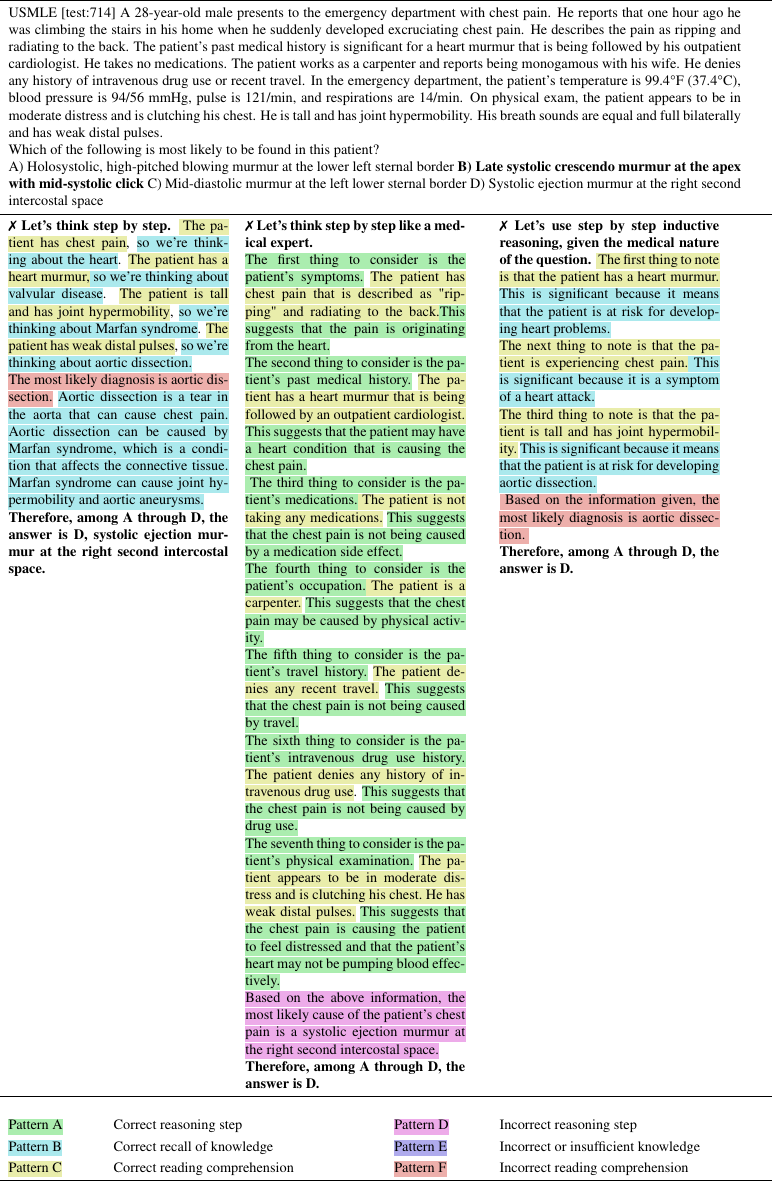}
\end{figure}

We provided nine more expert-labelled chain-of-thoughts in Figures \ref{fig:usmle-sample-2}, \ref{fig:usmle-sample-3}, \ref{fig:usmle-sample-4}, \ref{fig:usmle-sample-5}, \ref{fig:usmle-sample-6}, \ref{fig:usmle-sample-7}, \ref{fig:usmle-sample-8}, \ref{fig:usmle-sample-9} and \ref{fig:usmle-sample-10}. Note that patterns reported in Table \ref{tab:patterns} cannot always be matched to text segments, as one highlighted text segment does not always correspond to a single category (reasoning and knowledge patterns are often entangled).

\clearpage

\clearpage

\end{document}